\def\year{2020}\relax
\documentclass[letterpaper]{article} 
\usepackage{aaai20}  
\usepackage{times}  
\usepackage{helvet} 
\usepackage{courier}  
\usepackage[hyphens]{url}  
\usepackage{graphicx} 
\usepackage{graphicx}  
\usepackage{epsfig}
\usepackage{amsmath}
\usepackage{amssymb}
\usepackage[]{subfig}
\usepackage{tabularx}
\usepackage{enumitem}
\usepackage{booktabs}
\usepackage{makecell}
\usepackage{algorithm}
\usepackage{algorithmic}

\newcommand{\Commentl}[1]{ {\hfill $\triangleright$ \it #1}}
\newcommand{\Comment}[1]{ {\begin{flushright} $\triangleright$ \it #1 \end{flushright}}}

\title{Neural Architecture Search using Deep Neural Networks \\and Monte Carlo Tree Search}

\newcommand*\samethanks[1][\value{footnote}]{\footnotemark[#1]}

\author{Linnan Wang\thanks{Linnan Wang and Yiyang Zhao contributed equally}\textsuperscript{1} \quad
Yiyang Zhao\samethanks\textsuperscript{2} \quad
Yuu Jinnai\textsuperscript{1} \quad
Yuandong Tian\textsuperscript{3} \quad
Rodrigo Fonseca\textsuperscript{1} \quad\\
\textsuperscript{1}{Brown University} \quad
\textsuperscript{2}{Worcester Polytechnic Institute} \quad
\textsuperscript{3}{Facebook AI Research}}

\begin{document}
\maketitle

\begin{abstract}
Neural Architecture Search (NAS) has shown great success in automating the design of neural networks, but the prohibitive amount of computations behind current NAS methods requires further investigations in improving the sample efficiency and the network evaluation cost to get better results in a shorter time. In this paper, we present a novel scalable Monte Carlo Tree Search (MCTS) based NAS agent, named AlphaX, to tackle these two aspects. AlphaX improves the search efficiency by adaptively balancing the exploration and exploitation at the state level, and by a Meta-Deep Neural Network (DNN) to predict network accuracies for biasing the search toward a promising region. To amortize the network evaluation cost, AlphaX accelerates MCTS rollouts with a distributed design and reduces the number of epochs in evaluating a network by transfer learning, which is guided with the tree structure in MCTS. In 12 GPU days and 1000 samples, AlphaX found an architecture that reaches 97.84\% top-1 accuracy on CIFAR-10, and 75.5\% top-1 accuracy on ImageNet, exceeding SOTA NAS methods in both the accuracy and sampling efficiency. Particularly, we also evaluate AlphaX on NASBench-101, a large scale NAS dataset; AlphaX is 3x and 2.8x more sample efficient than Random Search and Regularized Evolution in finding the global optimum. Finally, we show the searched architecture improves a variety of vision applications from Neural Style Transfer, to Image Captioning and Object Detection. 
\end{abstract}

\section{Introduction}

Designing efficient neural architectures is extremely laborious. A typical design iteration starts with a heuristic design hypothesis from domain experts, followed by the design validation with hours of GPU training. The entire design process requires many of such iterations before finding a satisfying architecture. Neural Architecture Search has emerged as a promising tool to alleviate human effort in this trial and error design process, but the tremendous computing resources required by current NAS methods motivate us to investigate both the search efficiency and the network evaluation cost.

\begin{figure}[!htb]
\centering 
\subfloat[][random search]{\centering \includegraphics[height = 3.2cm]{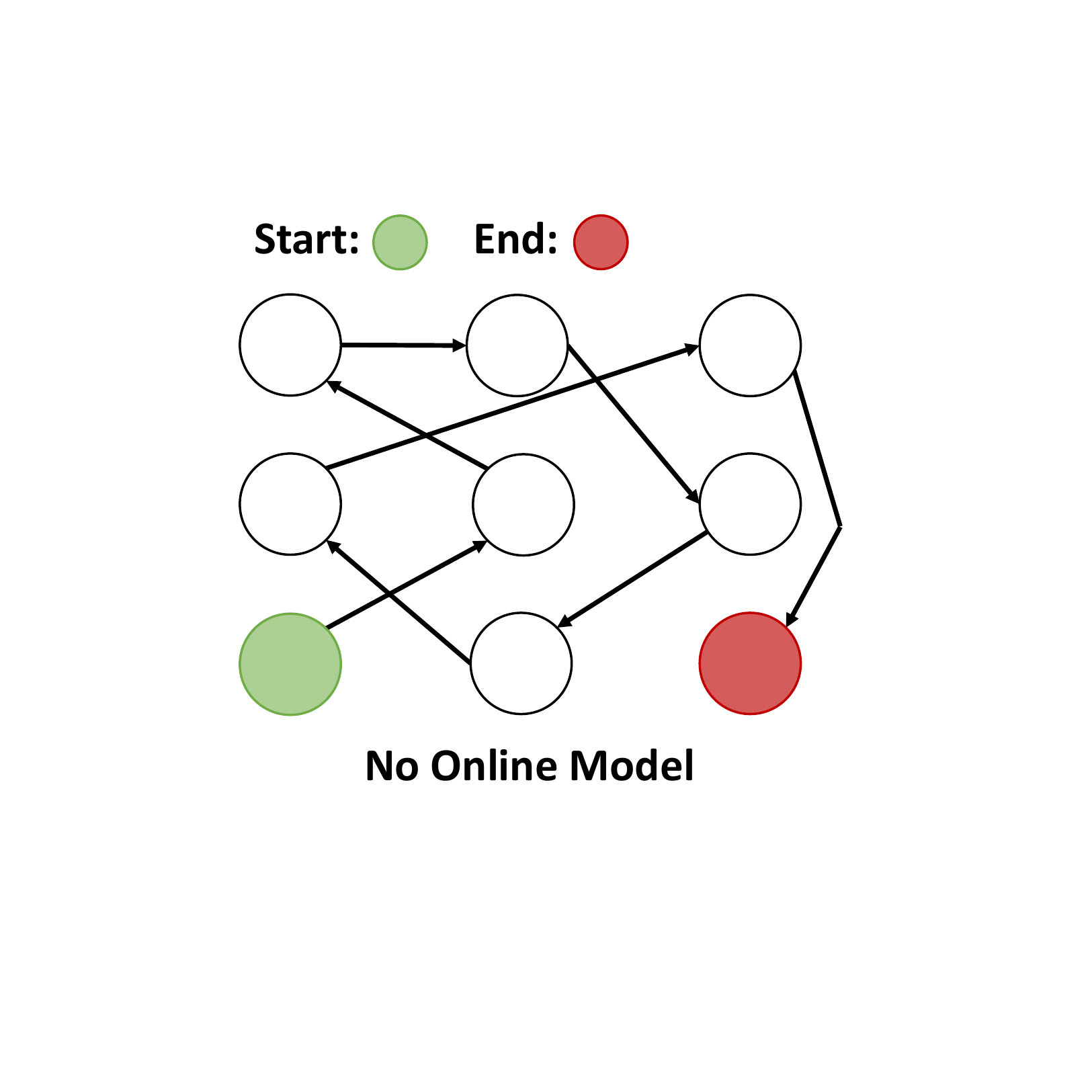}\label{design_domain}} \quad
\subfloat[][greedy methods]{\centering \includegraphics[height = 3.2cm]{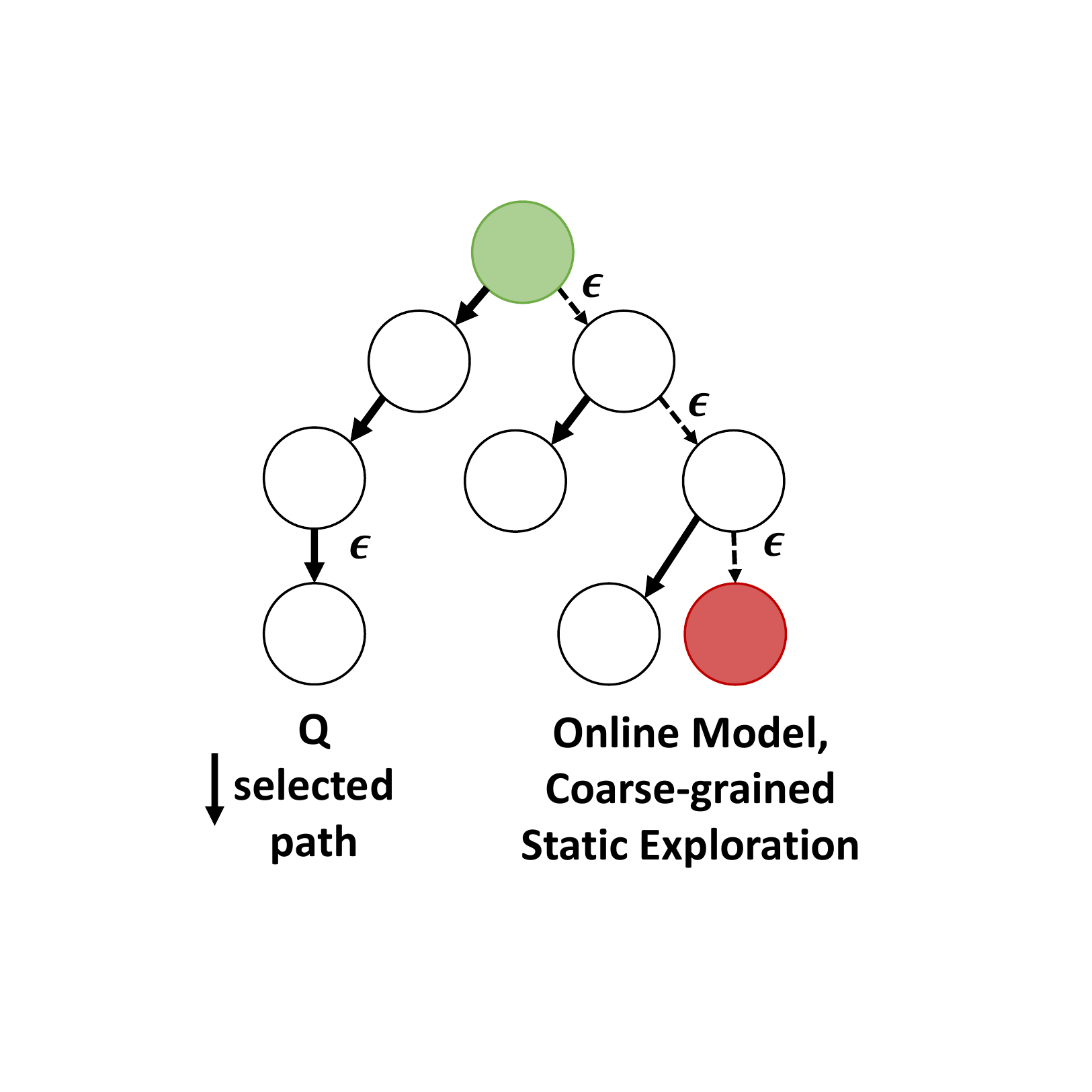}\label{MCTS_terminal_state}}\\
\subfloat[][MCTS]{\centering \includegraphics[height = 3.2cm]{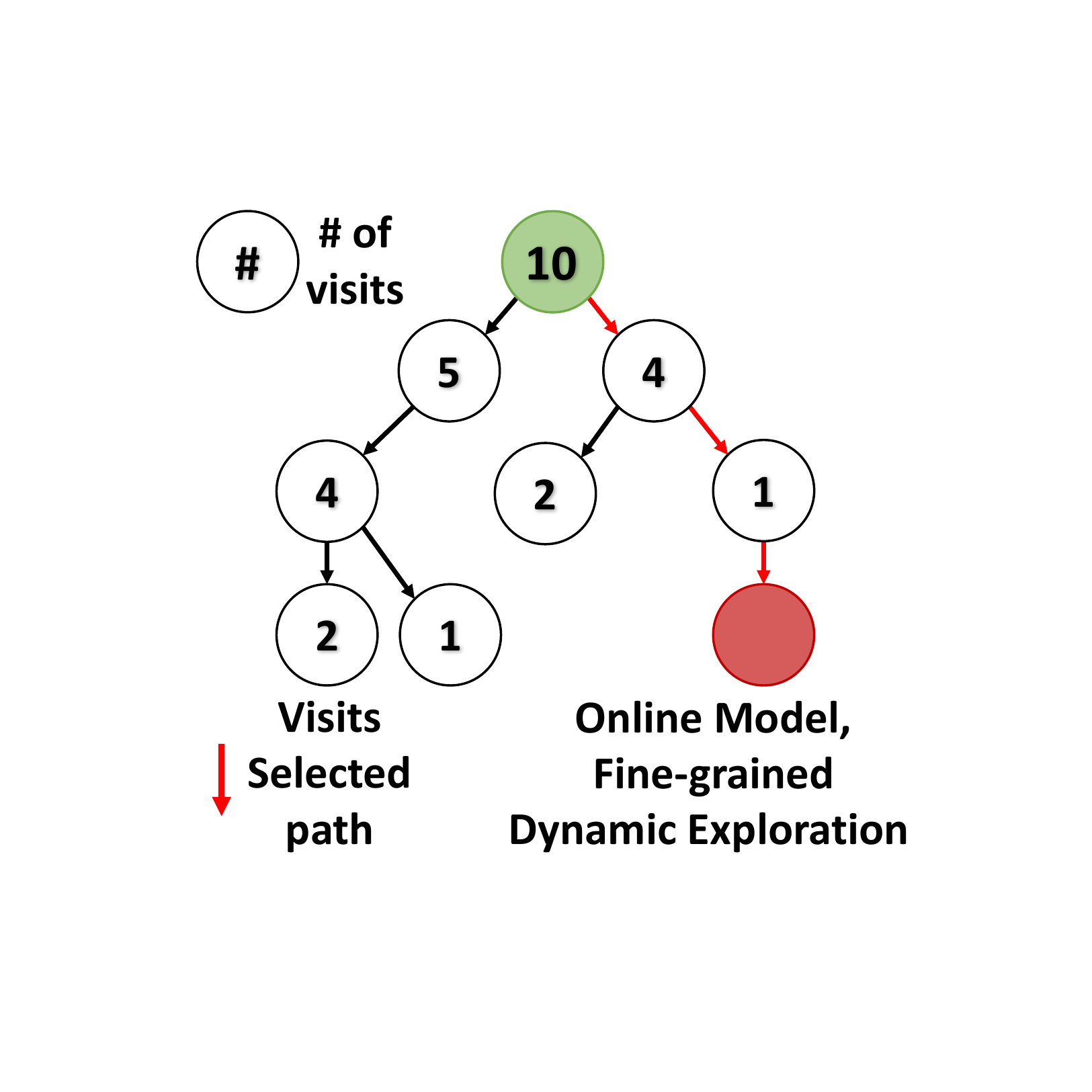}\label{MCTS_terminal_state}} \quad
\subfloat[][Performance]{\centering \includegraphics[height = 3.2cm]{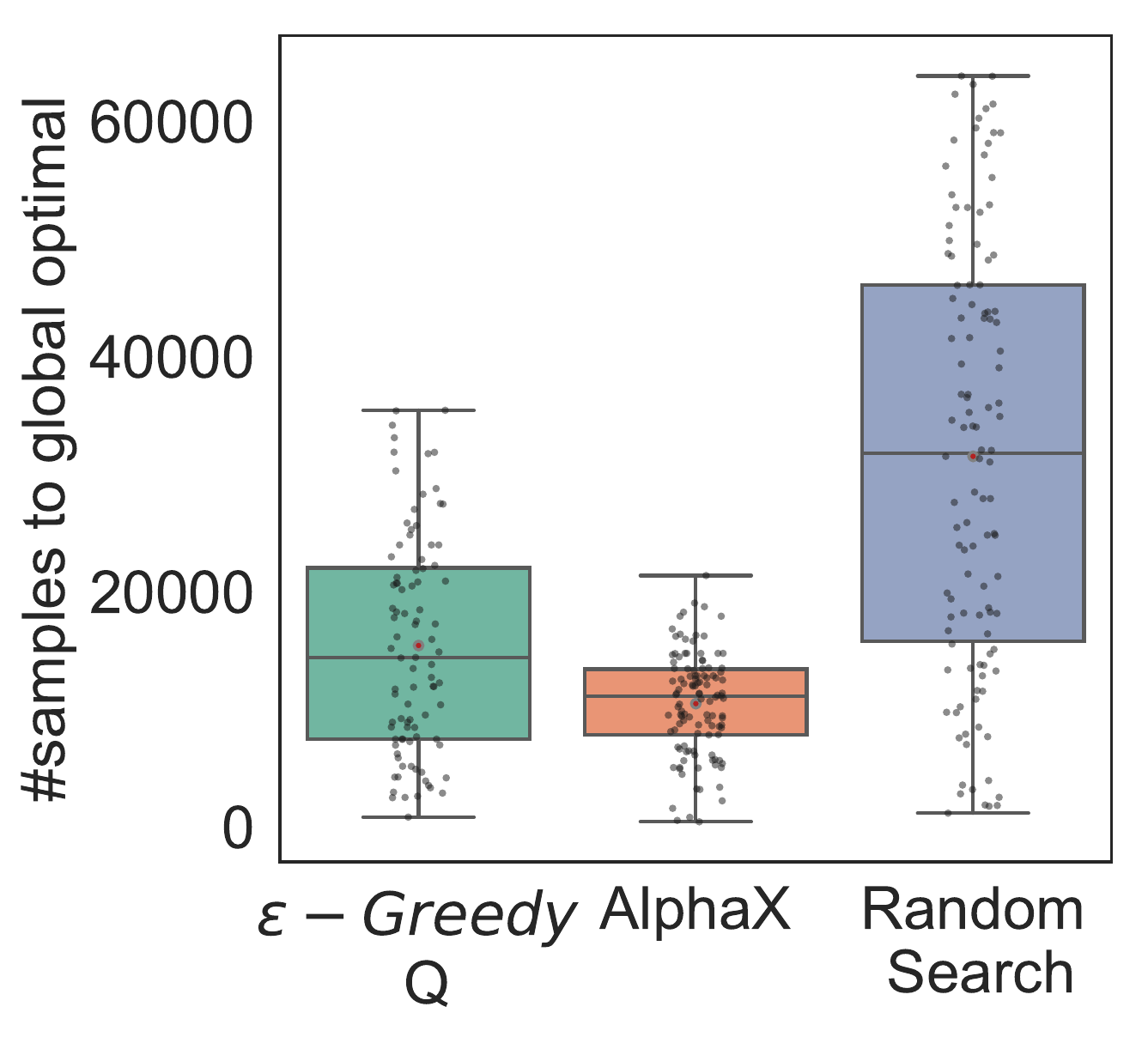}\label{mcts_speed_nasbench_teaser}}
\caption{ \textbf{Comparisons of NAS algorithms}: \textbf{(a)} \textit{random search} makes independent decision without using prior rollouts (previous search trajectories). An online model is to evaluate how promising the current search branch based on prior rollouts, and \textit{random search} has no online model. \textbf{(b)} Search methods guided by online performance models built from previous rollouts. With static, coarse-grained exploration strategy (e.g., $\epsilon$-greedy in \textit{Q-learning}), they may quickly be stuck in a sub-optimal solution; and the chance to escape is exponentially decreasing along the trajectory.  \textbf{(c)} AlphaX builds online models of both performance and visitation counts for adaptive exploration. The numbers in nodes represent values. \textbf{(d)} Performance of different search algorithms on NASBench-101. AlphaX is 3x, 1.5x more sample-efficient than \textit{random search} and \textit{$\epsilon$-greedy} based \textit{Q-learning}.  }
\label{fig:teaser}
\end{figure}

AlphaGo/AlphaGoZero~\cite{silver2016mastering,yuandong2019OpeGo} have recently shown super-human performance in playing the game of Go, by using a specific search algorithm called Monte-Carlo Tree Search (MCTS)~\cite{kocsis2006bandit}. Given the current game state, MCTS gradually builds an online model for its subsequent game states to evaluate the winning chance at that state, based on search experiences in the current and prior games, and makes a decision. The search experience is from the previous search trajectories (called \emph{rollouts}) that have been tried, and their consequences (whether the player wins or not). Different from the traditional MCTS approach that evaluates the consequence of a trajectory by random self-play to the end of a game, AlphaGo uses a \emph{predictive model} (or value network) to predict the consequence, which enjoys much lower variance. Furthermore, due to its built-in exploration mechanism using \textit{Upper Confidence bound applied to Trees} (UCT)~\cite{auer2002finite}, based on its online model, MCTS dynamically adapts itself to the most promising search regions, where good consequences are likely to happen.

Inspired by this idea, we present AlphaX, a NAS agent that uses MCTS for efficient architecture search with Meta-DNN as a predictive model to estimate the accuracy of a sampled architecture. Compared with Random Search, AlphaX builds an online model which guides the future search; compared to greedy methods, e.g. Q-learning, Regularized Evolution or Top-K methods, AlphaX dynamically trades off exploration and exploitation and can escape from locally optimal solutions with fewer search trials. Fig.~\ref{fig:teaser} summarizes the trade-offs. Furthermore, while prior works applied MCTS to Architecture Search~\cite{wistuba2017finding,negrinho2017deeparchitect}, they lack an effective model to accurately predict rewards, and the expensive network evaluations in MCTS rollouts still remain unaddressed. Toward a practical MCTS-based NAS agent, AlphaX has two novel features: first, a highly accurate multi-stage meta-DNN to improve the sample efficiency; and second, the use of transfer learning, together with a scalable distributed design, to amortize the network evaluation costs. As a result, AlphaX is the first MCTS-based NAS agent that reports SOTA accuracies on both CIFAR-10 and ImageNet in on par with the SOTA end-to-end search time.

\begin{table}[t]
\setlength{\tabcolsep}{0.1em}
  \scriptsize
  \label{layers_used}
  \centering
  \begin{tabular}{ l l l l l l }
    \toprule
          \textbf{Methods}     &\textbf{\thead{\scriptsize Global\\\scriptsize Solution }}     & \textbf{\thead{\scriptsize Online\\\scriptsize Model }}     & \textbf{Exploration} & \textbf{\thead{\scriptsize Distributed\\\scriptsize Ready }} & \textbf{\thead{\scriptsize Transfer\\\scriptsize Learning }} \\
    \midrule
      MetaQNN (QL) \footnotemark                                &  $\times$                     & -                 & {$\epsilon$-greedy}    & $\times$ & $\times$ \\
      Zoph (PG) \footnotemark                                           &  $\times$                     & RNN                      & -                      & $\surd$  & $\times$ \\
      PNAS (HC) \footnotemark                       &  $\times$                     & RNN                      & -                      & $\surd$  & $\times$ \\
      Regularized Evolution \footnotemark               &  $\surd$                      & -                        & top-k mutation         & $\surd$  & $\times$ \\
      Random Search \footnotemark                       &  $\surd$                      & -                        & random                 & $\surd$  & $\times$ \\
        DeepArchitect (MCTS) \footnotemark         &  $\surd$            & -                  & UCT                    & $\times$ & $\times$ \\ 
      Wistuba (MCTS) \footnotemark                   &  $\surd$                      & \thead{\scriptsize{Gaussian}}             & UCT      & $\times$ & $\times$ \\   
      AlphaX (MCTS)                       &  $\surd$                 & meta-DNN                 & UCT           & $\surd$  & $\surd$  \\      
    \bottomrule
    \label{acc-comps-imagenet}
  \end{tabular}
  \caption{\textbf{Highlights of NAS search algorithms}: compared to DeepArchitect and Wistuba, AlphaX makes several contributions toward a practical MCTS based NAS agent by improving the sample efficiency with an effective multi-stage metaDNN, and amortizes the network evaluation costs by a distirbuted design and transfer learning.}
  \vspace{-0.2in}
\end{table}

\footnotetext[1]{\cite{Baker2016}}
\footnotetext[2]{\cite{Zoph2016}} 
\footnotetext[3]{\cite{liu2017progressive}}
\footnotetext[4]{\cite{real2018regularized}}
\footnotetext[5]{\cite{sciuto2019evaluating}}  
\footnotetext[6]{\cite{negrinho2017deeparchitect}}
\footnotetext[7]{\cite{wistuba2017finding}}

\newpage

\section{Related Work}

\textit{Monte Carlo Tree Search:} DeepArchitect~\cite{negrinho2017deeparchitect} implemented vanilla MCTS for NAS without a predictive model, and Wistuba~\cite{wistuba2017finding} uses statistics from the current search (e.g., RAVE and Contextual Reward Prediction) to predict the performance of a state. In comparison, our performance estimate is from both searched rollouts so far and a model (meta-DNN) learned from performances of known architectures and can generalize to unseen architectures. Besides, the prior works do not address the expensive network evaluations in MCTS rollouts, while AlphaX solves this issue with a distributed design and transfer learning, making important improvements toward a practical MCTS based NAS agent.

\textit{Bayesian Optimization} (BO) is a popular method for the hyper-parameter search~\cite{kandasamy2015high}. BO has proven to be successful  
for small scale problems, e.g. hyper-parameter tuning of Stochastic Gradient Descent (SGD). Though BO variants, e.g. TPE~\cite{bergstra2011algorithms}, have tackled the cubic scaling issue in Gaussian Process, optimizing the acquisition function in a high dimensional search space that contains over $10^{17}$ neural architectures become the bottleneck. 

\textit{Reinforcement Learning (RL)}: Several RL techniques have been investigated for NAS~\cite{Baker2016,Zoph2016}.
Baker et al. proposed a Q-learning agent to design network architectures~\cite{Baker2016}. The agent takes a $\epsilon$-greedy policy: with probability $1-\epsilon$, it chooses the action that leads to the best expected return (i.e. accuracy) estimated by the current model, otherwise uniformly chooses an action. Zoph et al. built an RNN agent trained with Policy Gradient to design CNN and LSTM~\cite{Zoph2016}. However, directly maximizing the expected reward in Policy Gradient could lead to local optimal solution~\cite{nachum2016improving}.

\textit{Hill Climbing (HC)}: Elsken et al. proposed a simple hill climbing for NAS~\cite{elsken2017simple}. Starting from an architecture, they train every descendent network before moving to the best performing child. Liu et al. deployed a beam search which follows a similar procedure to hill climbing but selects the top-K architectures instead of only the best~\cite{liu2017progressive}. HC is akin to the vanilla Policy Gradient tending to trap into a local optimum from which it can never escape, while MCTS demonstrates provable convergence toward the global optimum given enough samples~\cite{kocsis2006bandit}.

\textit{Evolutionary Algorithm (EA)}: Evolutionary algorithms represent each neural network as a string of genes and search the architecture space by mutation and recombinations~\cite{real2018regularized}.
Strings which represent a neural network with top performance are selected to generate child models. The selection process is in lieu of exploitation, and the mutation is to encourage exploration. Still, GA algorithms do not consider the visiting statistics at individual states, and it lacks an online model to inform decisions.

\begin{figure*}
\centering 
\includegraphics[height=6cm]{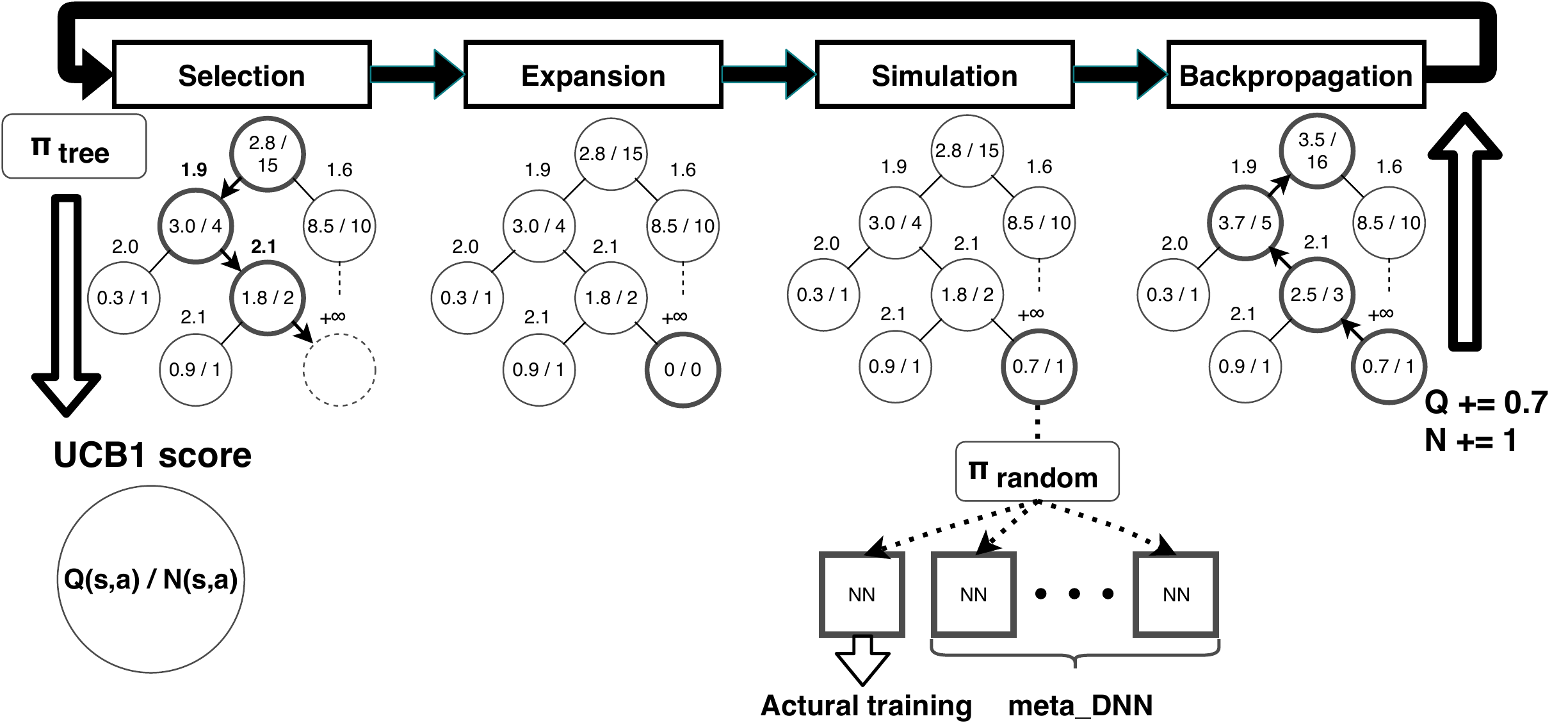}
\caption{\textbf{Overview of AlphaX search procedures:} explanations of four steps are in sec.\ref{sec:search}.}
\label{fig:mcts-procedures}
\end{figure*}

\begin{figure}[t]
\centering 
\subfloat[][NASNet Cell]{\includegraphics[height=1.1in]{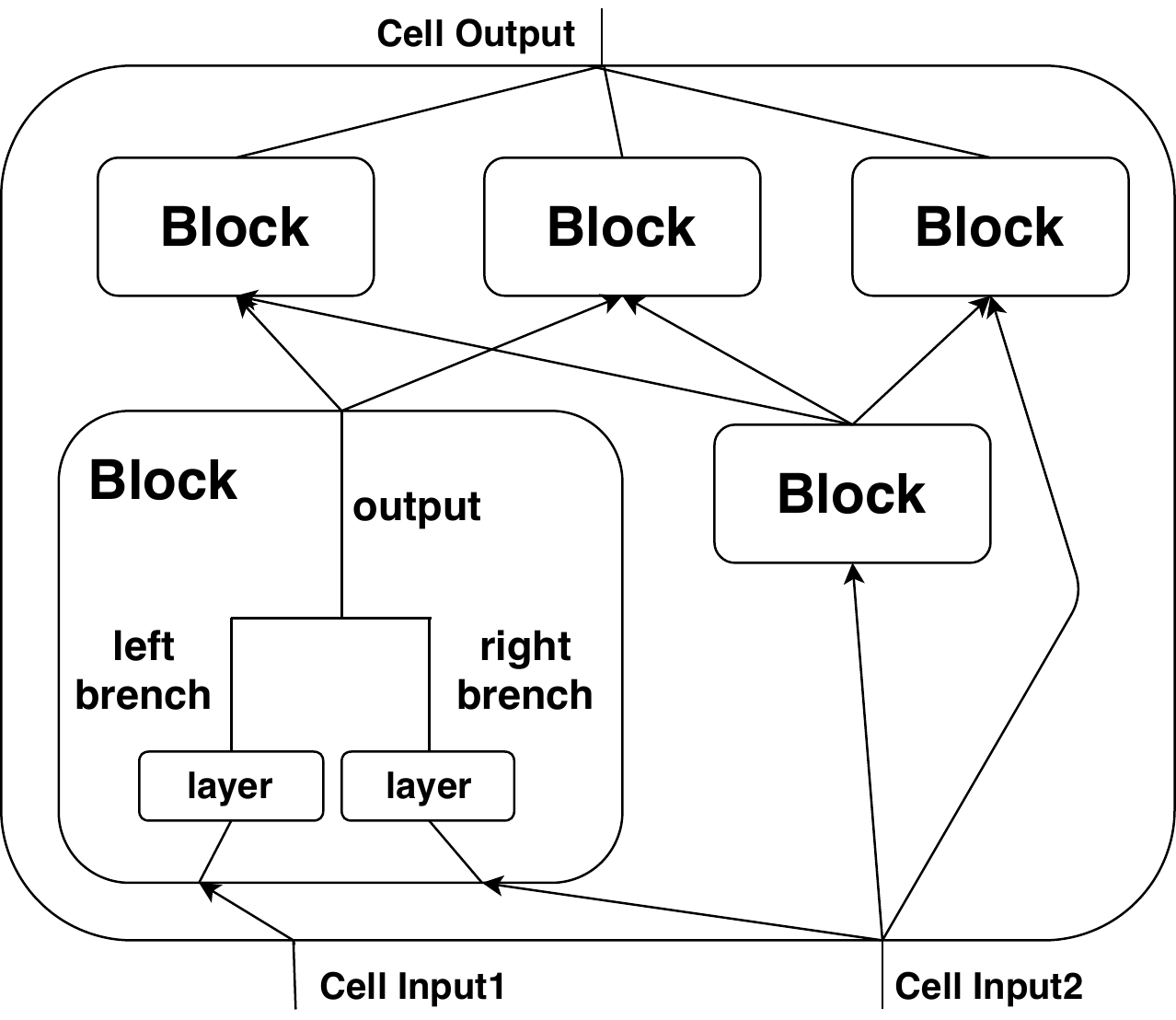}\label{design_domain}} \quad
\subfloat[][NASBench DAG]{\includegraphics[height=1.1in]{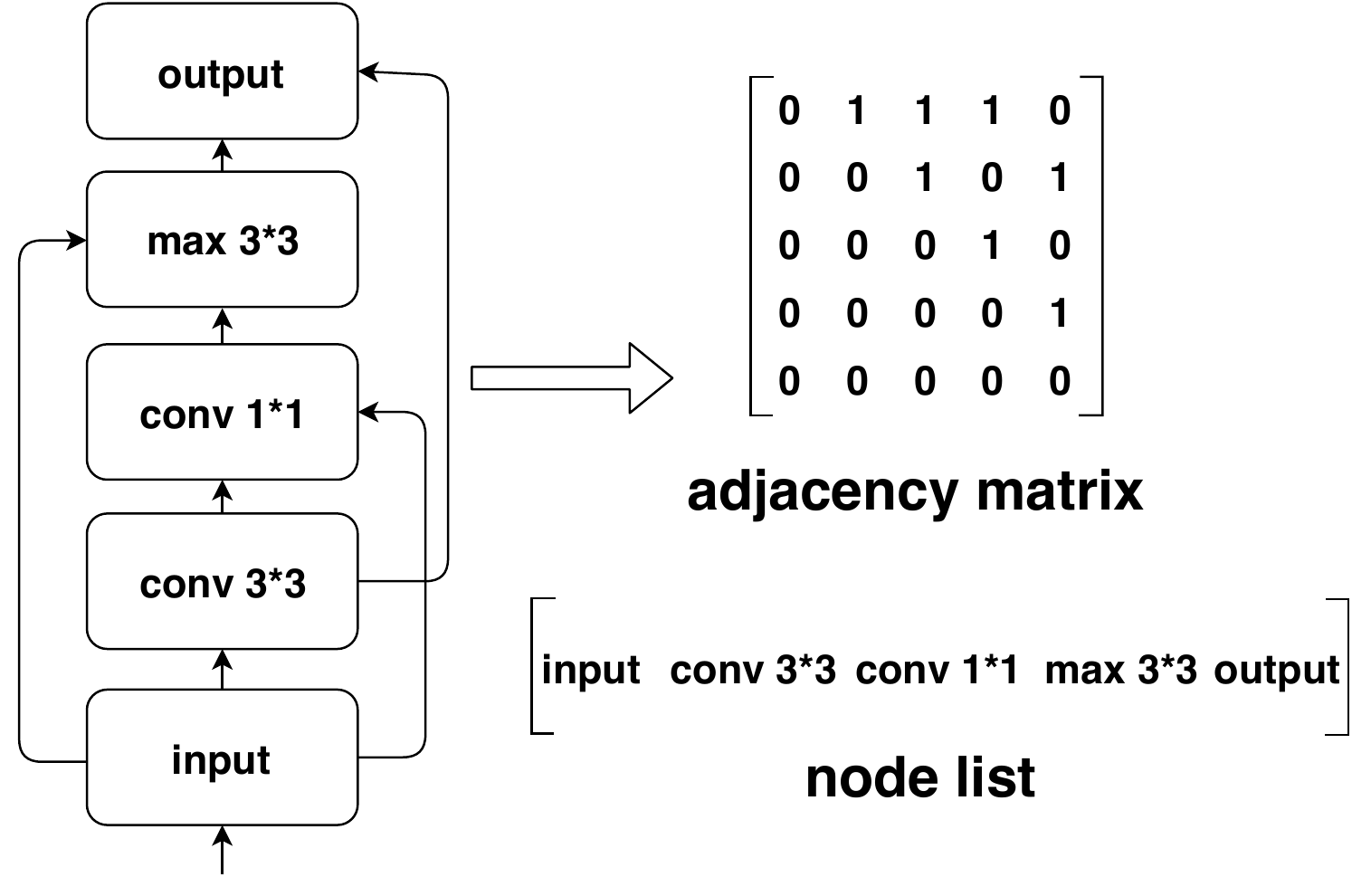}\label{MCTS_terminal_state}} 
\caption{ \textbf{Design space}: (a) the cell structure of NASNet and (b) the DAG structure of NASBench-101. Then the network is constructed by stacking multiple cells or DAGs. }
\label{nonlinear_connections}
\end{figure}

\section{AlphaX: A Scalable MCTS Design Agent}

\subsection{Design, State and Action Space}
\label{content:design_space}
\textbf{Design Space}: the neural architectures for different domain tasks, e.g. object detection and image classification, follow fundamentally different designs. This renders different design spaces for the design agent. AlphaX is flexible to support various search spaces with an intuitive state and action abstraction. Here we provide a brief description of two search spaces used in our experiments.  

\begin{itemize}
\item \textit{NASNet Search Space}: \cite{zoph2017learning} proposes searching a hierarchical Cell structure as shown in Fig.\ref{design_domain}. There are two types of Cells, Normal Cell ($NCell$) and Reduction Cell ($RCell$). $NCell$ maintains the input and output dimensions with the padding, while $RCell$ reduces the height and width by half with the striding. Then, the network is constituted by stacking multiple cells. 

\item \textit{NASBench Search Space}: \cite{ying2019bench} proposes searching a small Direct Acyclic Graph (DAG) with each node representing a layer and each edge representing the inter-layer dependency as shown in Fig.\ref{MCTS_terminal_state}. Similarly, the network is constituted by stacking multiple such DAGs. 
\end{itemize}

\label{sec:state}

\textbf{State Space}: a state represents a network architecture, and AlphaX utilizes states (or nodes) to keep track of past trails to inform future decisions. We implement a state as a map that defines all the hyper-parameters for each network layer and their dependencies. We also introduce a special terminal state to allow for multiple actions. All the other states can transit to the terminal state by taking the terminal action, and the agent only trains the network, from which it reaches the terminal. With the terminal state, the agent freely modifies the architecture before reaching the terminal. This enables multiple actions for the design agent to bypass shallow architectures.

\textbf{Action Space}: an action morphs the current network architecture, i.e. current state, to transit to the next state. It not only explicitly specifies the inter-layer connectivity, but also all the necessary hyper-parameters for each layer. Unlike games, actions in NAS are dynamically changing w.r.t the current state and design spaces. For example, AlphaX needs to leverage the current DAG (state) in enumerating all the feasible actions of 'adding an edge'. In our experiments, the actions for the NASNet search domain are adding a new layer in the left or right branch of a Block in a cell, creating a new block with different input combinations, and the terminating action. The actions for the NASBench search domain are either adding a node or an edge, and the terminating action.

\subsection{Search Procedure}

\label{sec:search}

This section elaborates the integration of MCTS and metaDNN. The purpose of MCTS is to analyze the most promising move at a state, while the purpose of meta-DNN is to learn the sampled architecture performance and to generalize to unexplored architectures so that MCTS can simulate many rollouts with only an actual training in evaluating a new node. The superior search efficiency of AlphaX is due to balancing the exploration and exploitation at the finest granularity, i.e. state level, by leveraging the visiting statistics. Each node tracks these two statistics: 1) $N(s, a)$ counts the selection of action $a$ at state $s$; 2) $Q(s, a)$ is the expected reward after taking action $a$ at state $s$, and intuitively $Q(s, a)$ is an estimate of how promising this search direction is. Fig.\ref{fig:mcts-procedures} demonstrates a typical searching iteration in AlphaX, which consists of \textit{Selection}, \textit{Expansion}, \textit{Meta-DNN assisted Simulation}, and \textit{Backpropagation}. We elucidate each step as follows.

\textbf{\textit{Selection}} traverses down the search tree to trace the current most promising search path. It starts from the root and stops till reaching a leaf. At a node, the agent selects actions based on UCB1 \cite{auer2002finite}:
\begin{equation}
	\label{ucb1_update}
	\pi_{tree}(s) = \text{arg}\max_{a \in A} \left( \frac{Q(s, a)}{N(s, a)} + 2 c \sqrt{\frac{2 \log N(s)}{N(s, a)}} \right),
\end{equation}
where $N(s)$ is the number of visits to the state $s$ (i.e. $N(s) = \sum_{a \in A} N(s, a)$), and $c$ is a constant. The first term ($\frac{Q(s, a)}{N(s, a)} $) is the exploitation term estimating the expected accuracy of its descendants. The second term ($2 c \sqrt{\frac{2 \log N(s)}{N(s, a)}}$) is the exploration term encouraging less visited nodes. The exploration term dominates $\pi_{tree}(s)$ if $N(s, a)$ is small, and the exploitation term otherwise. As a result, the agent favors the exploration in the beginning until building proper confidences to exploit. $c$ controls the weight of exploration, and it is empirically set to 0.5. We iterate the tree policy to reach a new node.

\textbf{\textit{Expansion}} adds a new node into the tree. $Q(s, a)$ and $N(s, a)$ are initialized to zeros. $Q(s, a)$ will be updated in the simulation step.

\textbf{\textit{Meta-DNN assisted Simulation}} randomly samples the descendants of a new node to approximate $Q(s, a)$ of the node with their accuracies.
The process is to estimate how promising the search direction rendered by the new node and its descendants. The simulation starts at the new node. The agent traverses down the tree by taking the uniform-random action until reaching a terminal state, then it dispatches the architecture for training.

The more simulation we roll, the more accurate estimate of this search direction we get. However, we cannot conduct many simulations as network training is extremely time-consuming.
AlphaX adopts a novel hybrid strategy to solve this issue by incorporating a meta-DNN to predict the network accuracy in addition to the actual training. We delay the introduction of meta-DNN to sec.\ref{sec:metadnn}.
Specifically, we estimate $q = Q(s, a)$ with
\begin{equation}
	\label{reward}
	Q(s, a) \leftarrow \left( Acc(sim_0(s')) + \frac{1}{k} \sum_{i=1..k} Pred(sim_i(s')) \right) / 2
\end{equation}
where $s' = s + a$, and $sim(s')$ represents a simulation starting from state $s'$. $Acc$ is the actually trained accuracy in the first simulation, and $Pred$ is the predicted accuracy from Meta-DNN in subsequent $k$ simulations. If a search branch renders architectures similar to previously trained good ones, Meta-DNN updates the exploitation term in Eq.\ref{ucb1_update} to increase the likelihood of going to this branch.

\textbf{\textit{Backpropagation}} back-tracks the search path from the new node to the root to update visiting statistics. Please note we discuss the sequential case here, and the backpropagation will be split into two parts in the distributed setting (sec.\ref{sec:distributed}). 
With the estimated $q$ for the new node, we iteratively back-propagate the information to its ancestral as:
\begin{equation}
	\begin{split}
	& Q(s, a) \leftarrow Q(s, a) + q, \quad
	 N(s, a) \leftarrow N(s, a) + 1 \quad \\
	& s \leftarrow parent(s), \quad
	 a \leftarrow \pi_{tree}(s)
	\end{split}
\end{equation}
until it reaches the root node.

\begin{figure}
\centering 
\includegraphics[height=1.7in]{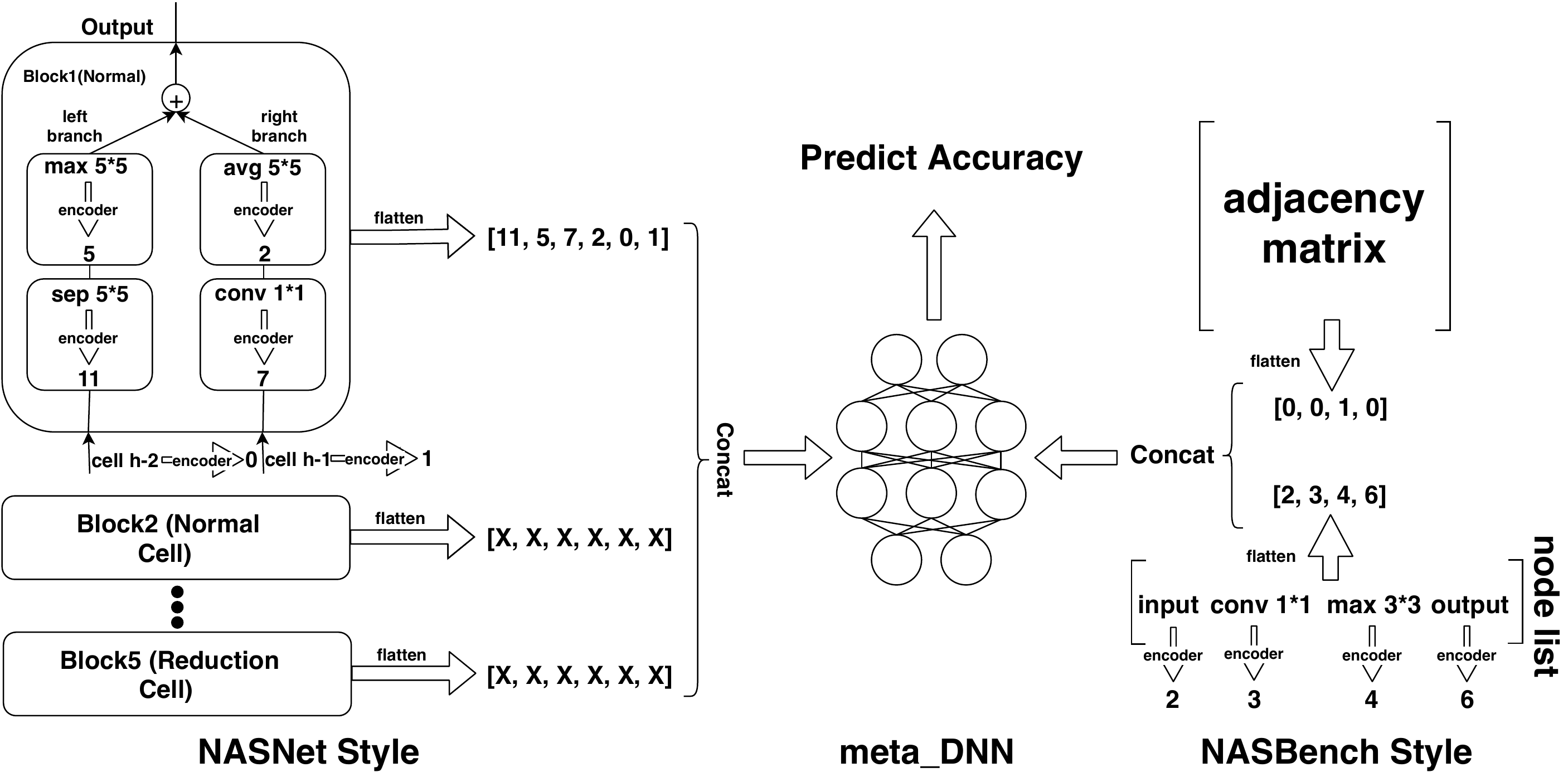}
\caption{Encoding scheme of NASBench and NASNet.}
\label{fig:network-encoding}
\end{figure}

\subsection{The design of Meta-DNN and its related issues}
\label{sec:metadnn}

Meta-DNN intends to generalize the performance of unseen architectures based on previously sampled networks. It provides a practical solution to accurately estimate a search branch with many simulations without involving the actual training (see the metaDNN assisted simulation for details). New training data is generated as AlphaX advances in the search. So, the learning of Meta-DNN is end-to-end. The input of Meta-DNN is a vector representation of architecture, while the output is the prediction of architecture performance, i.e. test accuracy. 

The coding scheme for NASNet architectures is as follows: we use 6-digits vector to code a $Block$; the first two digits represent up to two layers in the left branch, and the 3rd and 4th digits for the right branch. Each layer is represented by a number in $[1, 12]$ to represent 12 different layers, and the specific layer code is available in Appendix TABLE.\ref{app:layers_used}. We use 0 to pad the vector if a layer is absent. The last two digits represent the input for the left and right branch, respectively. For the coding of block inputs, 0 corresponds to the output of the previous $Cell$, 1 is the previous, previous $Cell$, and $i+2$ is the output of $Block_i$. If a block is absent, it is [0,0,0,0,0,0]. The left part of Fig.\ref{fig:network-encoding} demonstrates an example of NASNet encoding scheme. A $Cell$ has up to 5 blocks, so a vector of 60 digits is sufficient to represent a state that fully specifies both $RCell$ and $NCell$. The coding scheme for NASBench architectures is a vector of flat adjacency matrix, plus the nodelist. Similarly, we pad 0 if a layer or an edge is absent. The right part of Fig.\ref{fig:network-encoding} demonstrates an example of NASBench encoding scheme. Since NASBench limits nodes $\leq$ 7, 7$\times$7 (adjacency matrix)+ 7 (nodelist) = 56 digits can fully specify a NASBench architecture.

Now we cast the prediction of architecture performance as a regression problem. Finding a good metaDNN is heuristically oriented and it should vary from tasks to tasks. We calculate the correlation between predicted accuracies and true accuracies from the sampled architectures in evaluating the design of metaDNN. Ideally, the metaDNN is expected to rank an unseen architecture in roughly similar to its true test accuracy, i.e. corr = 1. Various ML models, such as Gaussian Process, Neural Networks, or Decision Tree, are candidates for this regression task. We choose Neural Networks as the backbone model for its powerful generalization on the high-dimensional data and the online training capability. More ablations studies for the specific choices of metaDNN are available in sec.\ref{sec:metadnn_design}.

\subsection{ Transfer Learning }
\begin{figure}
    \begin{center}
      \includegraphics[height=3.8cm]{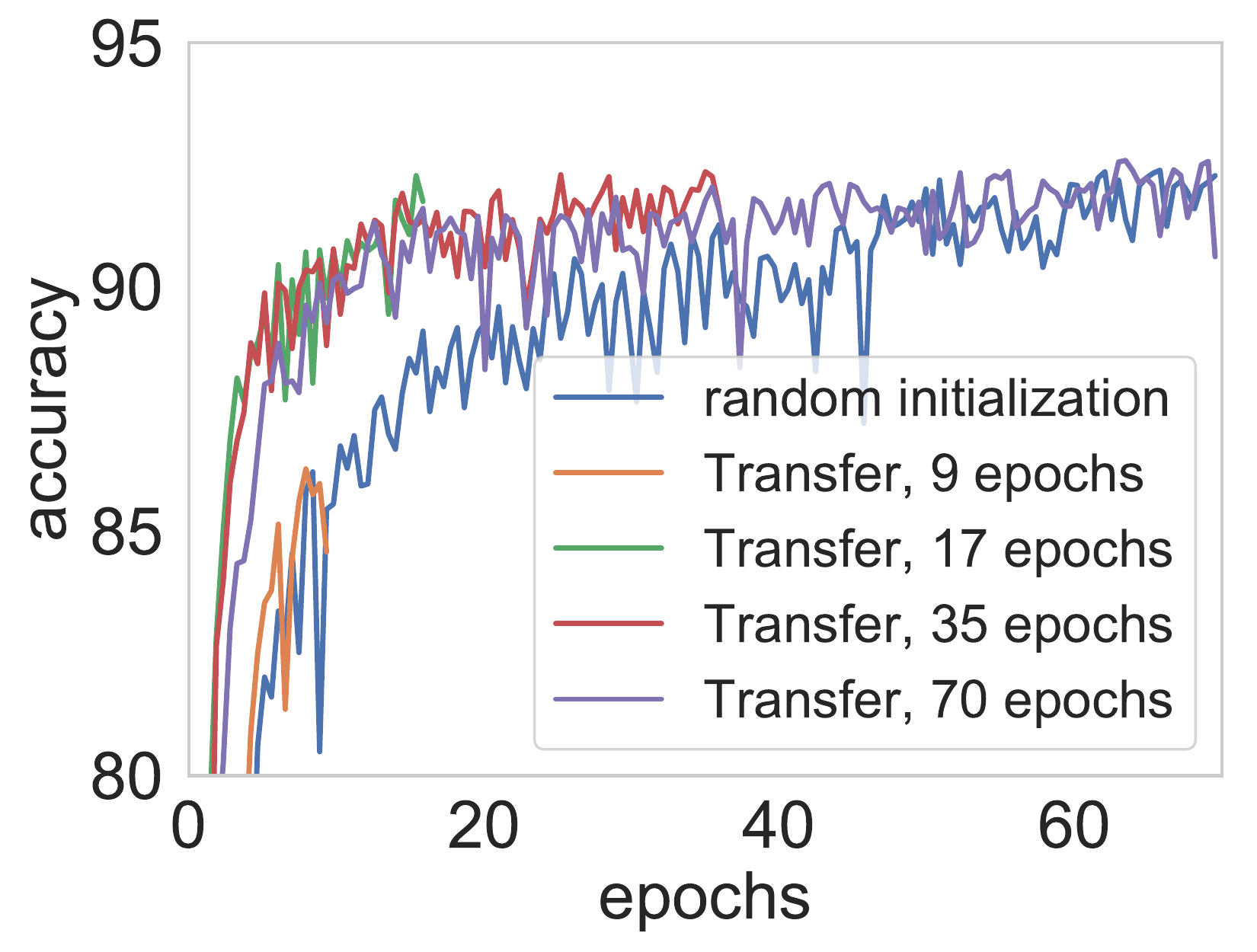}
    \end{center}
    \caption{\textbf{Validation of transfer learning}: transferring weights significantly reduces the number of epochs in reaching the same accuracy of random initializations (Transfer $17\rightarrow70$ epochs v.s. random initialization), but insufficient epochs loses accuracy (Transfer, 9 epochs).}
    \label{fig:transfer_learning}
  \end{figure}

As MCTS incrementally builds a network with primitive actions, networks falling on the same search path render similar structures. This motivates us to incorporate transfer learning in AlphaX to speed network evaluations up. In simulation (Fig.~\ref{fig:mcts-procedures}), AlphaX recursively traverses up the tree to find a previously trained network with the minimal edit distance to the newly sampled network. Then we transfer the weights of overlapping layers, and randomly initialize new layers. In the pre-training, we train every sample for 70 epochs if no parent networks are transferable, and 20 epochs otherwise. Fig.~\ref{fig:transfer_learning} provides a study to justify the design.

\subsection{Distributed AlphaX}
\label{sec:distributed}

\begin{figure}
  \begin{center}
    \includegraphics[width=0.85\columnwidth]{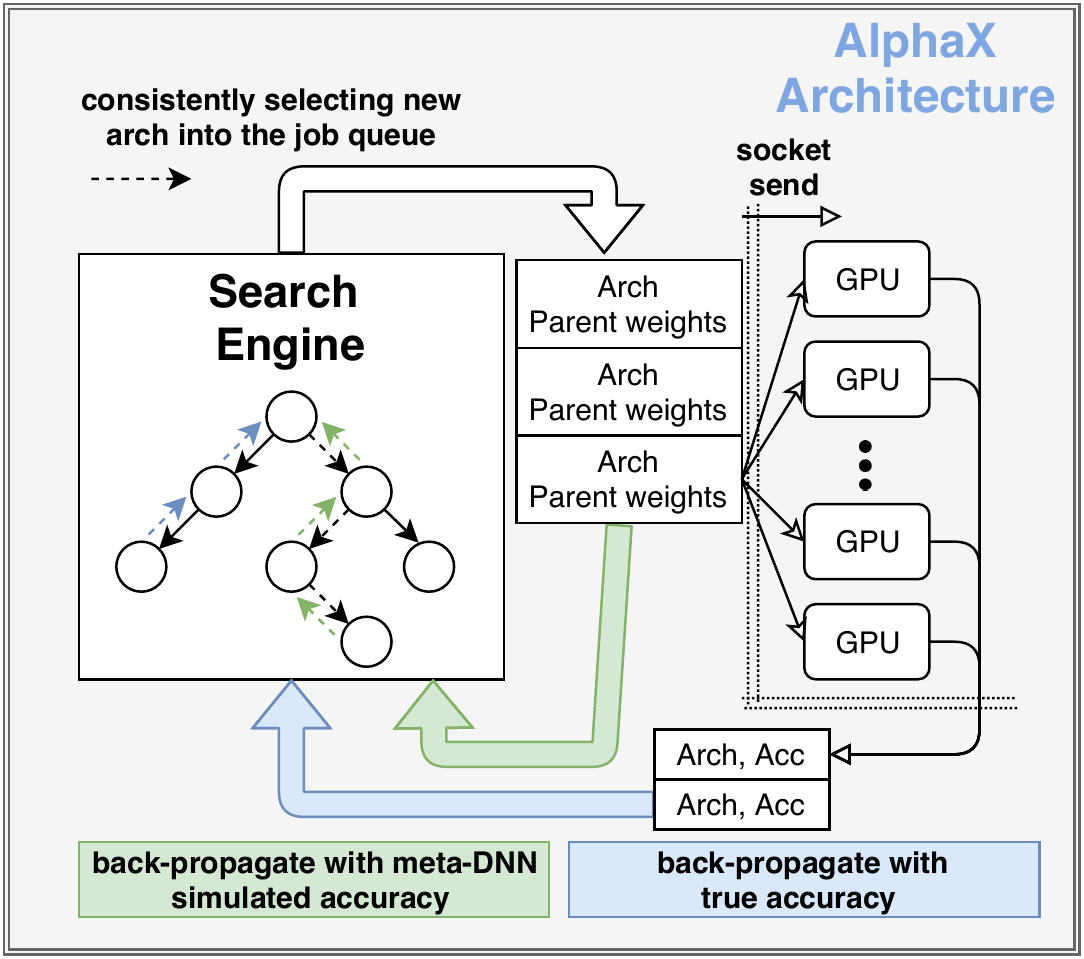}
  \end{center}
  \caption{\textbf{Distributed AlphaX}: we decouple the original back-propagation into two parts: one uses predicted accuracy (green arrow), while the other uses the true accuracy (blue arrow).  The pseudocode for the whole system is available in Appendix Sec.\ref{ap:pseudocode}  }
  \vspace{-0.1in}
  \label{distributed_alphaX}
\end{figure}

It is imperative to parallelize AlphaX to work on a large scale distributed systems to tackle the computation challenges rendered by NAS.
Fig.\ref{distributed_alphaX} demonstrates the distributed AlphaX. There is a master node exclusively for scheduling the search, while there are multiple clients (GPU) exclusively for training networks. The general procedures on the server side are as follows: 1) The agent follows the selection and expansion steps described in Fig.\ref{fig:mcts-procedures}. 2) The simulation in MCTS picks a network $arch_n$ for the actual training, and the agent traverses back to find the weights of parent architecture having the minimal edit distance to $arch_n$ for transfer learning; then we push both $arch_n$ and parent weights into a job queue. We define $arch_n$ as the selected network architecture at iteration $n$, and $rollout_{-}from(arch_n)$ as the node which it started the rollout from to reach $arch_n$. 3) The agent {\it preemptively backpropagates} $\hat{q} \leftarrow \frac{1}{k} \sum_{i=1..k} Pred(sim_i(s'))$ based only on predicted accuracies from the Meta-DNN at iteration $n$.
\begin{equation}
	\begin{split}
	&Q(s, a) \leftarrow Q(s, a) + \hat{q}, \quad
	N(s, a) \leftarrow N(s, a) + 1, \quad \\
	&s \leftarrow parent(s), \quad
	a \leftarrow \pi_{tree}(s).
	\end{split}
\end{equation}
4) The server checks the receive buffer to retrieve a finished job from clients that includes $arch_z$, $acc_z$.
Then the agent starts the second backpropagation to propagate $q \leftarrow \frac{acc_z + \hat{q}}{2}$ (Eq. \ref{reward}) from the node the rollout started ($s \leftarrow rollout_{-}from(arch_z)$) to replace the backpropagated $\hat{q}$ with $q$:

\begin{equation}
	\begin{split}
	Q(s, a) \leftarrow Q(s, a) + q - \hat{q}, \quad \\
	s \leftarrow parent(s), \quad 
	a \leftarrow \pi_{tree}(s).
	\end{split}
\end{equation}

The client constantly tries to retrieve a job from the master job queue if it is free. It starts training once it gets the job, then it transmits the finished job back to the server. So, each client is a dedicated trainer. We also consider the fault-tolerance by taking a snapshot of the server's states every few iterations, and AlphaX can resume the searching from the breakpoint using the latest snapshot.

\begin{figure}
  \begin{center}
	  \subfloat[][normal cell]{\centering \includegraphics[width=0.6\columnwidth]{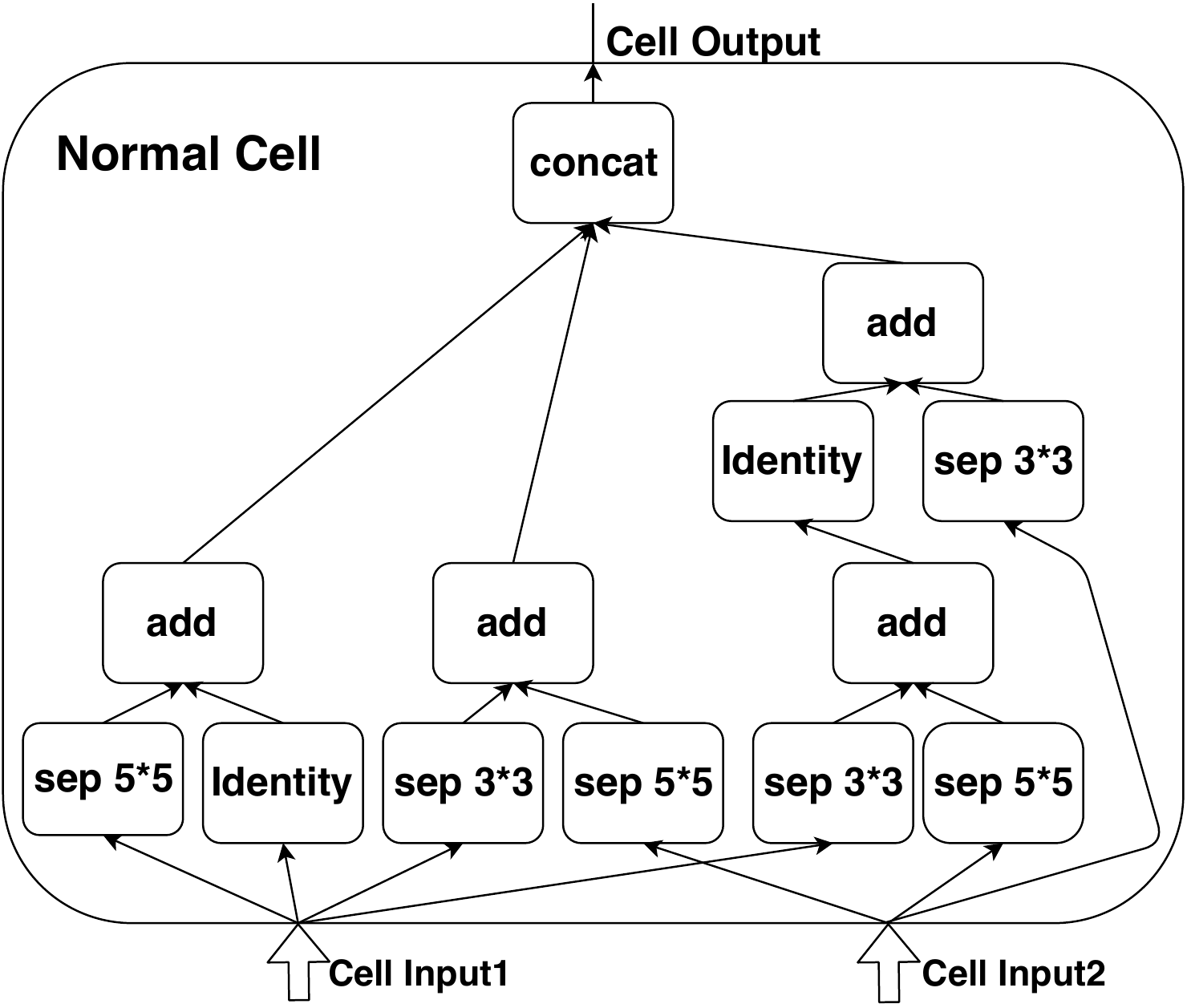}}          \\
	  \subfloat[][reduction cell]{\centering \includegraphics[width=0.6\columnwidth]{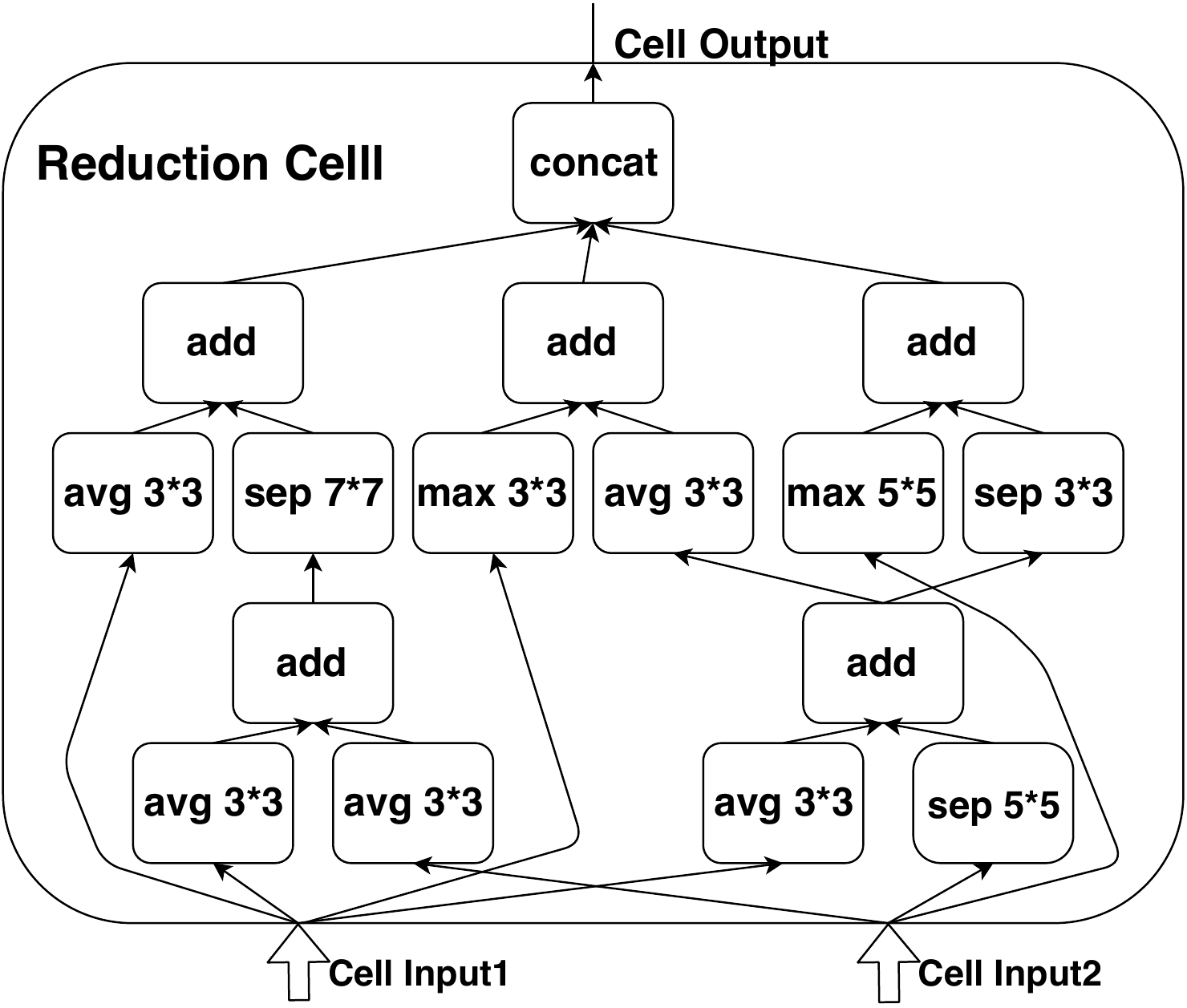}}	  
  \end{center}
  \caption{the normal and reduction cells that yield the highest accuracy in the search.}
  \label{best-arch}
\end{figure}

\begin{table}[t]
\setlength{\tabcolsep}{0.1em}
  \scriptsize
  \centering
  \begin{tabular}{l l l l l l}
    \toprule
         \textbf{Model}                                     & \textbf{Params}  & \textbf{Err} & \textbf{GPU days} & \textbf{M}\\
    \midrule
		  NASNet-A+cutout    \cite{zoph2017learning}        & 3.3M   & 2.65             & 2000  & 20000  \\
		  AmoebaNet-B+cutout \cite{real2018regularized}     & 2.8M   & $2.50_{\pm0.05}$ & 3150  & 27000  \\
  		  DARTS+cutout       \cite{liu2018darts}            & 3.3M   & $2.76_{\pm0.09}$ & 4     & -      \\
		  RENASNet+cutout    \cite{chen2019renas}           & 3.5M   & $2.88_{\pm0.02}$ & 6     & 4500   \\
		  \midrule
		  AlphaX+cutout (32 filters)  		            	& 2.83M  & $2.54_{\pm0.06}$ & 12    & 1000   \\
		  \midrule
		  PNAS               \cite{liu2017progressive}      & 3.2M   & $3.41_{\pm0.09}$ & 225   & 1160   \\
		  ENAS               \cite{pham2018efficient}       & 4.6M   & 3.54             & 0.45  &  -     \\
		  NAONet             \cite{luo2018neural}           & 10.6M  & 3.18             & 200   & 1000   \\
		  \midrule
          AlphaX (32 filters)                               & 2.83M & $3.04_{\pm0.03}$  & 12    & 1000   \\
	\midrule
  		  NAS v3\cite{Zoph2016}				                & 7.1M   & 4.47             & 22400 & 12800  \\ 
          Hier-EA	  \cite{liu2017hierarchical}            & 15.7M  & $3.75_{\pm0.12}$ & 300   & 7000   \\
	\midrule
		  AlphaX+cutout (128 filters)                       & 31.36M  & $2.16_{\pm0.04}$ & 12    & 1000   \\		  
    \bottomrule
  \end{tabular}
  \caption{The comparisons of our NASNet search results to other state-of-the-art results on CIFAR-10. M is the number of sampled architectures in the search. The cell structure of AlphaX is in Fig.~\ref{best-arch}.}
  \label{acc-comps-cifar}
  \footnotetext{* represents cutout}
\end{table}

\begin{table}[t]
\setlength{\tabcolsep}{0.2em}
  \scriptsize
  \centering
  \begin{tabular}{ l l l l l l}
    \toprule
          \textbf{model} 							& \textbf{multi-adds}      & \textbf{params} & \textbf{top1/top5 err}  \\
    \midrule
		  NASNet-A      \cite{zoph2017learning}     &    564M				   & 5.3M   & 26.0/8.4 \\
		  AmoebaNet-B   \cite{real2018regularized}  &    555M                  & 5.3M   & 26.0/8.5 \\
		  DARTS         \cite{liu2018darts}         &    574M                  & 4.7M   & 26.7/8.7 \\
		  RENASNet      \cite{chen2019renas}        &    574M                  & 4.7M   & 24.3/7.4 \\
		  PNAS          \cite{liu2017progressive}   &    588M                  & 5.1M   & 25.8/8.1 \\
	\midrule
		  AlphaX-1  						        &    579M                  & 5.4M   & 24.5/7.8 \\
    \bottomrule
  \end{tabular}
  \caption{ Transferring the CIFAR architecture to ImageNet, and their results in the mobile setting. }
  \label{acc-comps-imagenet}
\end{table}

  \begin{figure}[t]
    \begin{center}
      \subfloat[][best acc progression]{ \includegraphics[height=3.3cm]{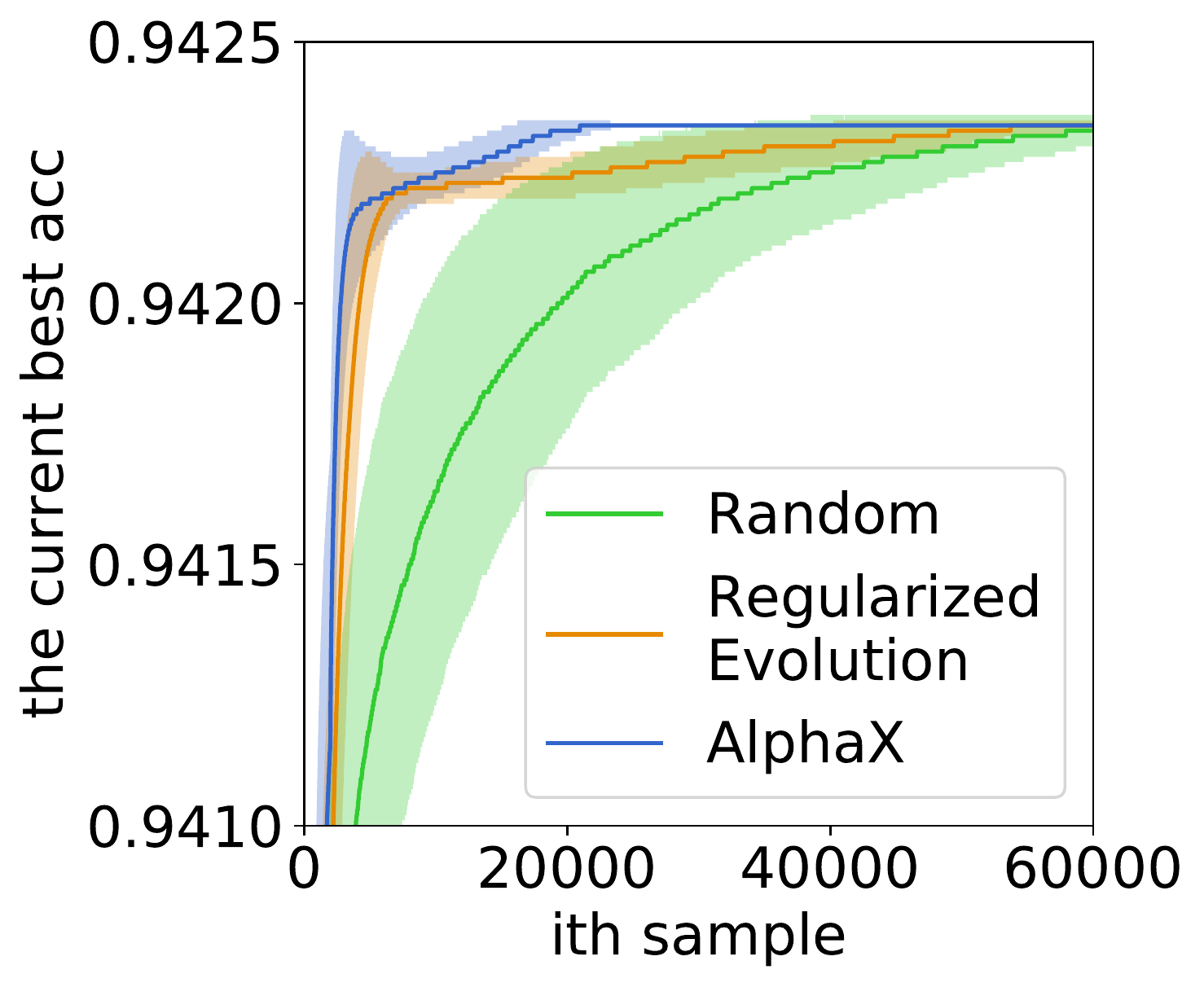}\label{fig:nasbench_best_trace}}   \quad
      \subfloat[][\#samples to the best]{ \includegraphics[height=3.3cm]{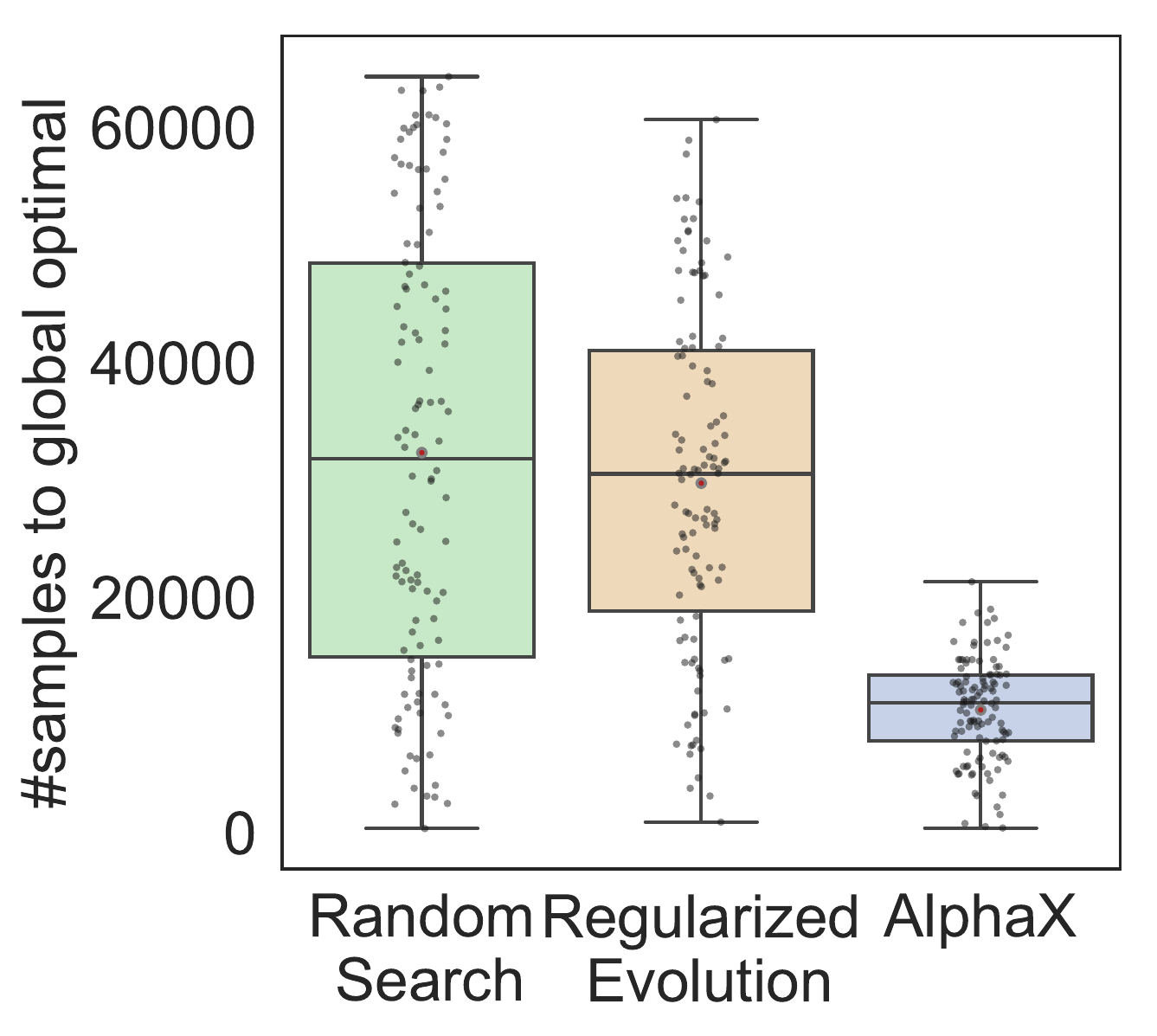}\label{fig:nasbench_samples_to_best}} \quad
      \subfloat[][best acc progression]{ \includegraphics[height=3.3cm]{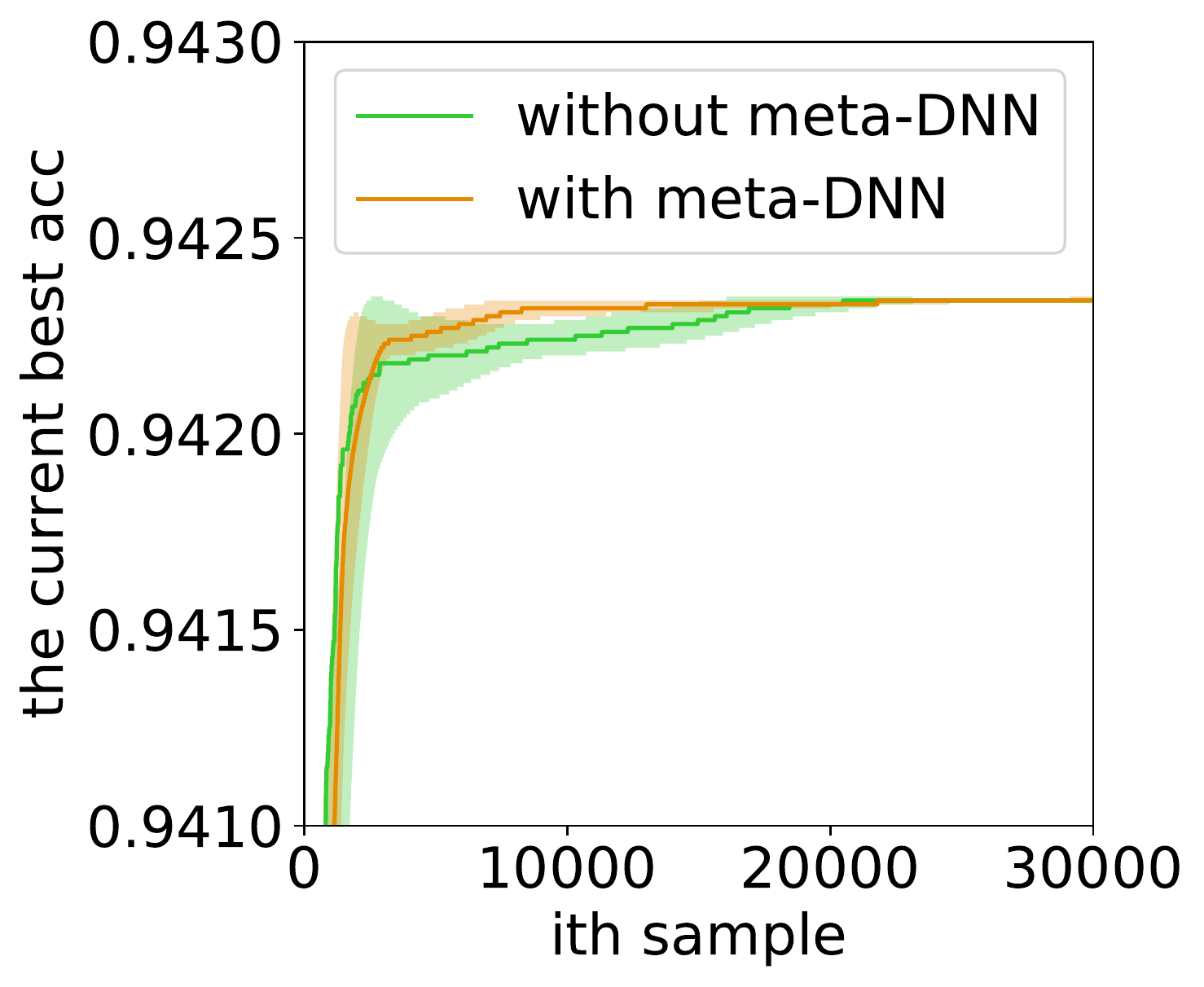}\label{fig:nasbench-metadnn-best-trace}}   \quad
      \subfloat[][\#samples to the best]{ \includegraphics[height=3.3cm]{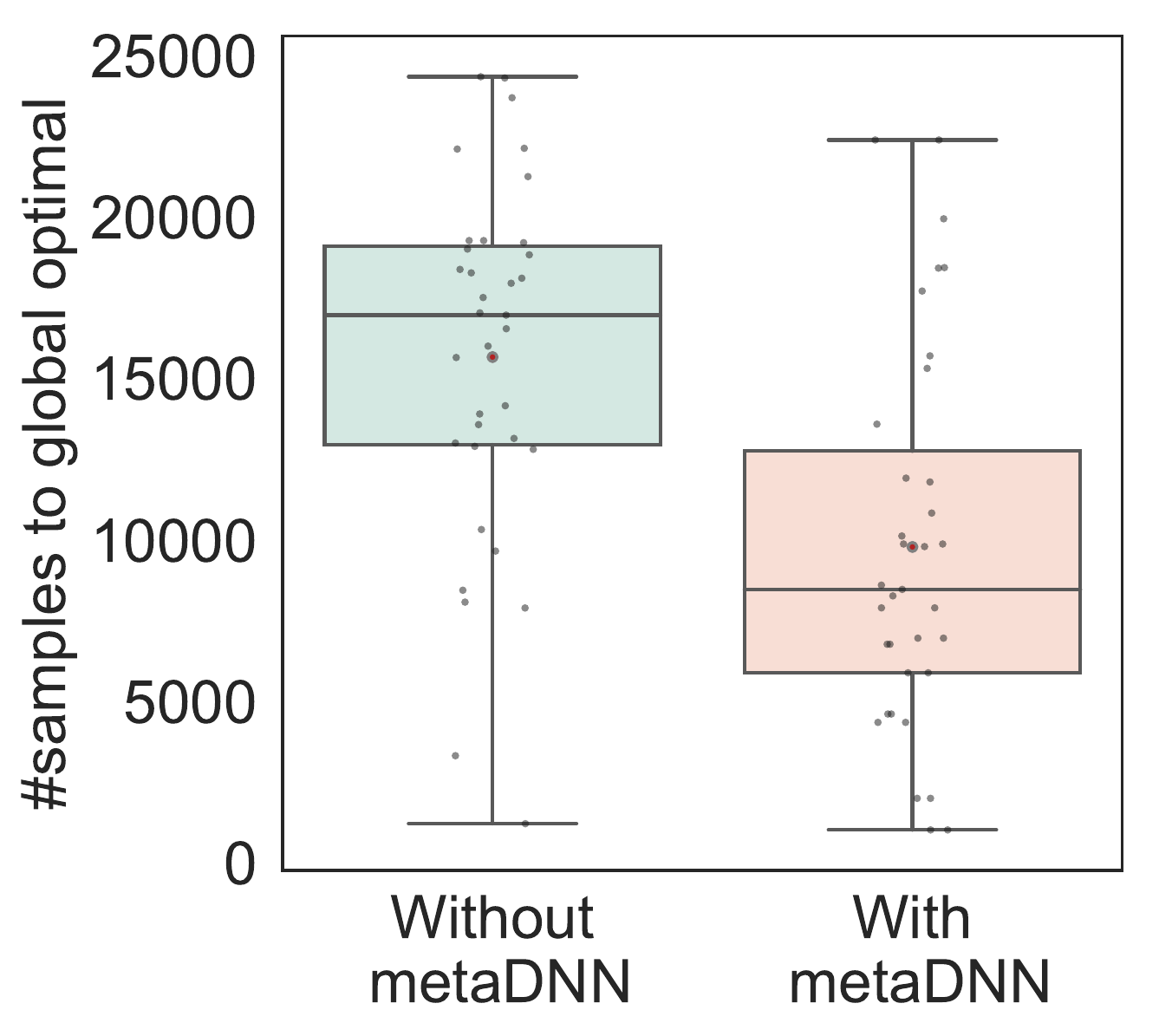}\label{fig:nasbench-metadnn-samples-to-best}}
    \end{center}
    \caption{\textbf{Finding the global optimum on NASBench-101}: AlphaX is 3x, 2.8x faster than Random Search and Regularized Evolution on NASBench-101 (nodes $\leq 6$). The results are from 200 trails with different random seeds. 
    (c) and (d) show the performance of AlphaX in cases of with/without meta-DNN on NASBench-101}
    \label{fig:nasbench-results}
  \end{figure}
  
  \begin{figure}
    \begin{center}
      \includegraphics[width=0.8\columnwidth ]{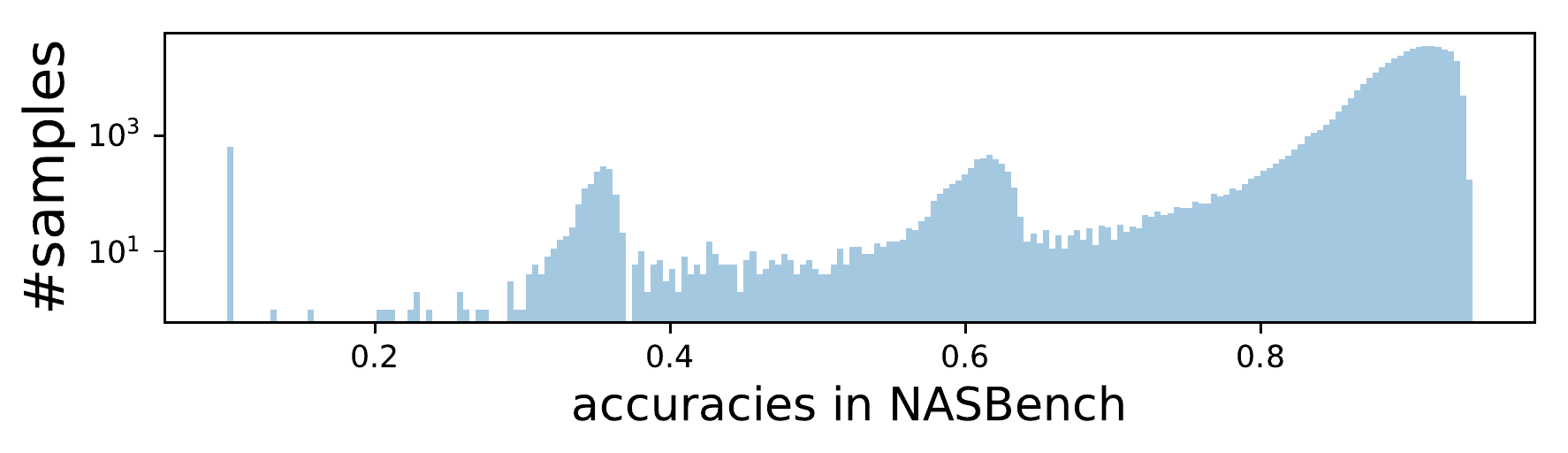}
    \end{center}
    \caption{Accuracy distribution of NASBench.}
    \label{fig:nasbench_acc_dist}
  \end{figure}
  
  \begin{figure}[t]
  	\begin{center}
  	  \includegraphics[width=0.8\columnwidth, trim={1cm 0cm 1cm 0cm}]{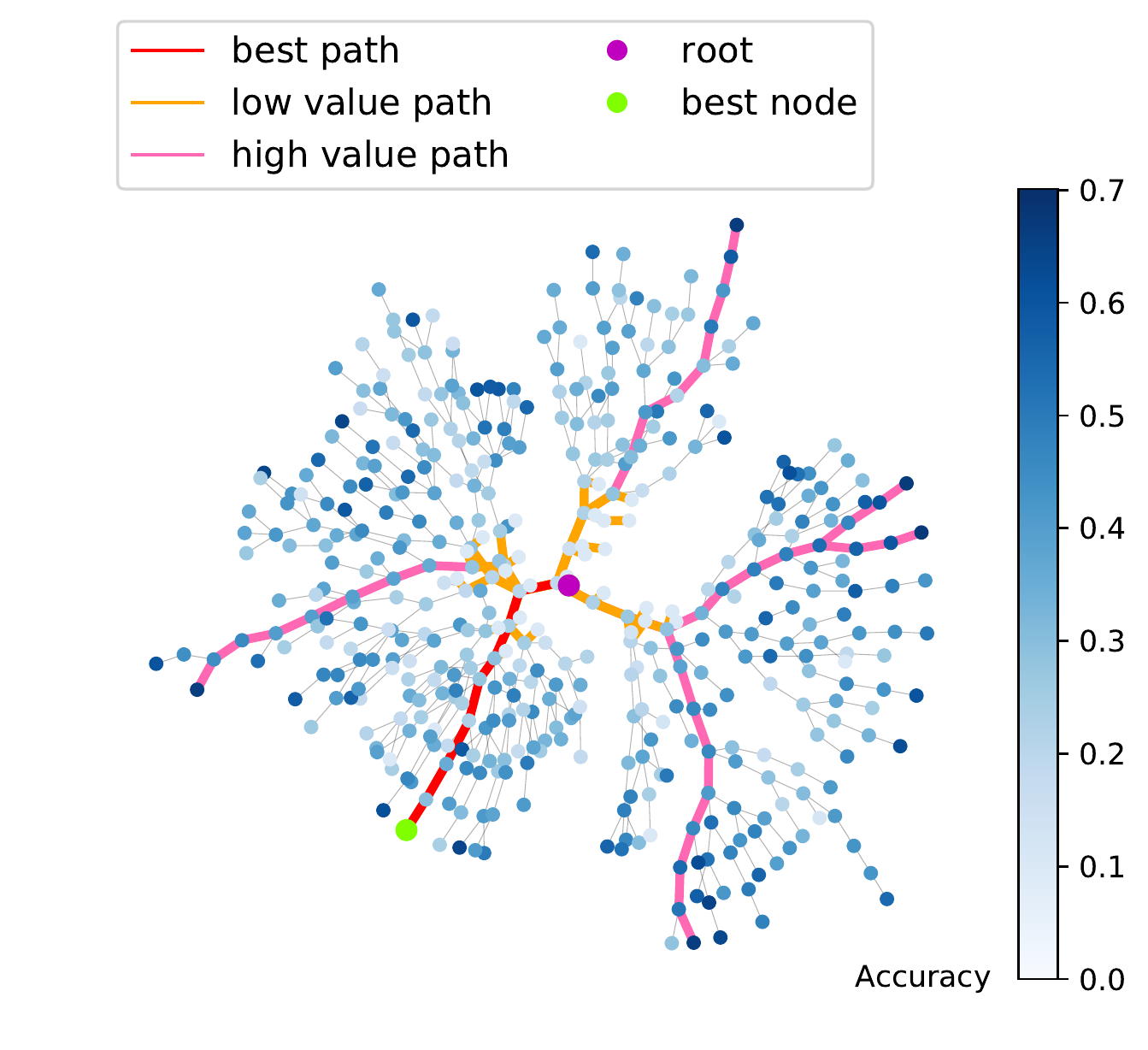}
  	\end{center}
  	\caption{\textbf{AlphaX search visualization}:each node represents a MCTS state; the node color reflects its value, i.e. accuracy, indicating how promising a search branch.  }
  	\vspace{-0.1in}
  	\label{fig:mcts_viz}
  \end{figure}
  
  \begin{figure}[t]
  \centering
  \subfloat[][RNN train]{ \includegraphics[width=0.49\columnwidth]{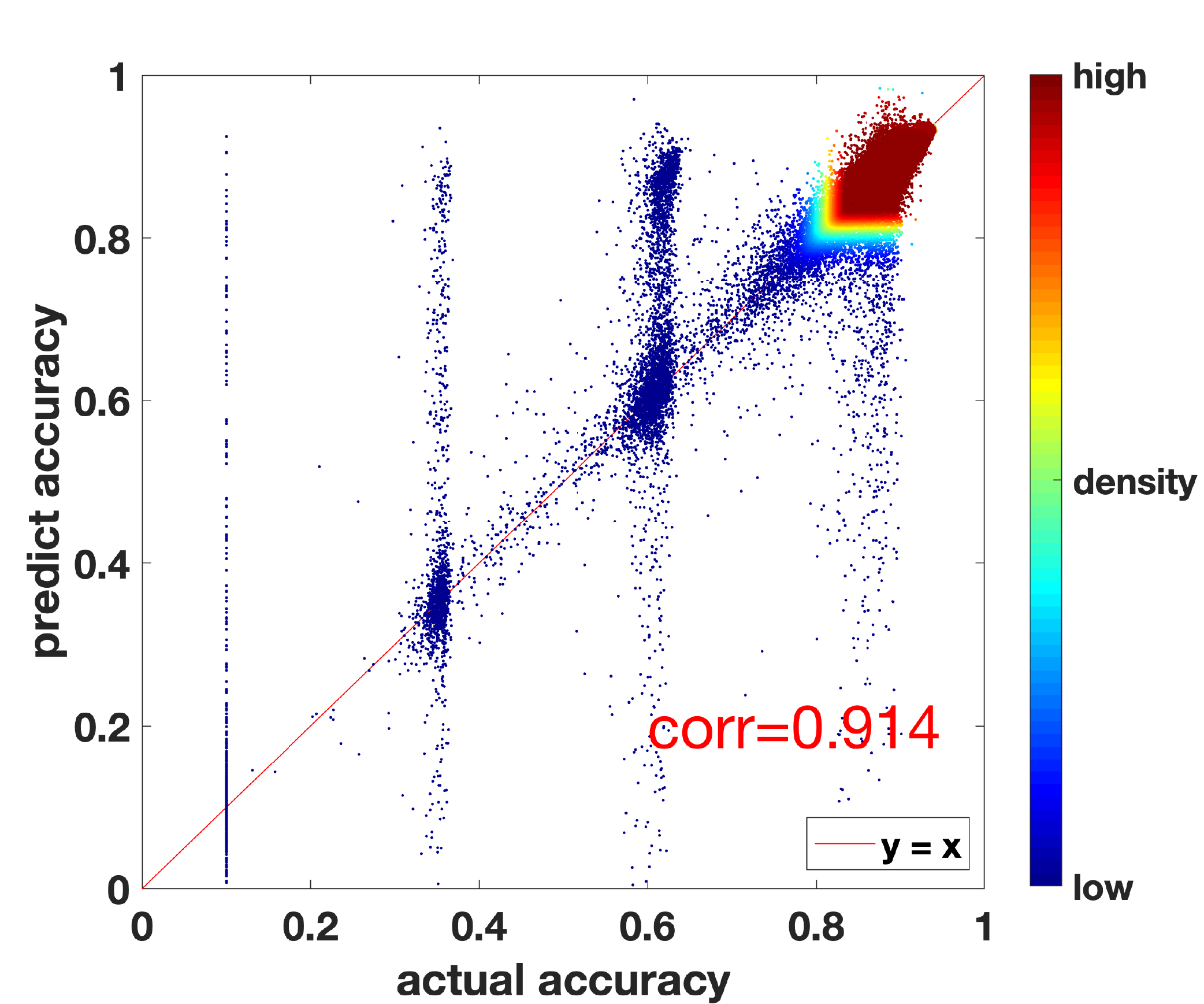}\label{fig:metadnn-rnn-train}}
  \subfloat[][RNN test]{ \includegraphics[width=0.49\columnwidth]{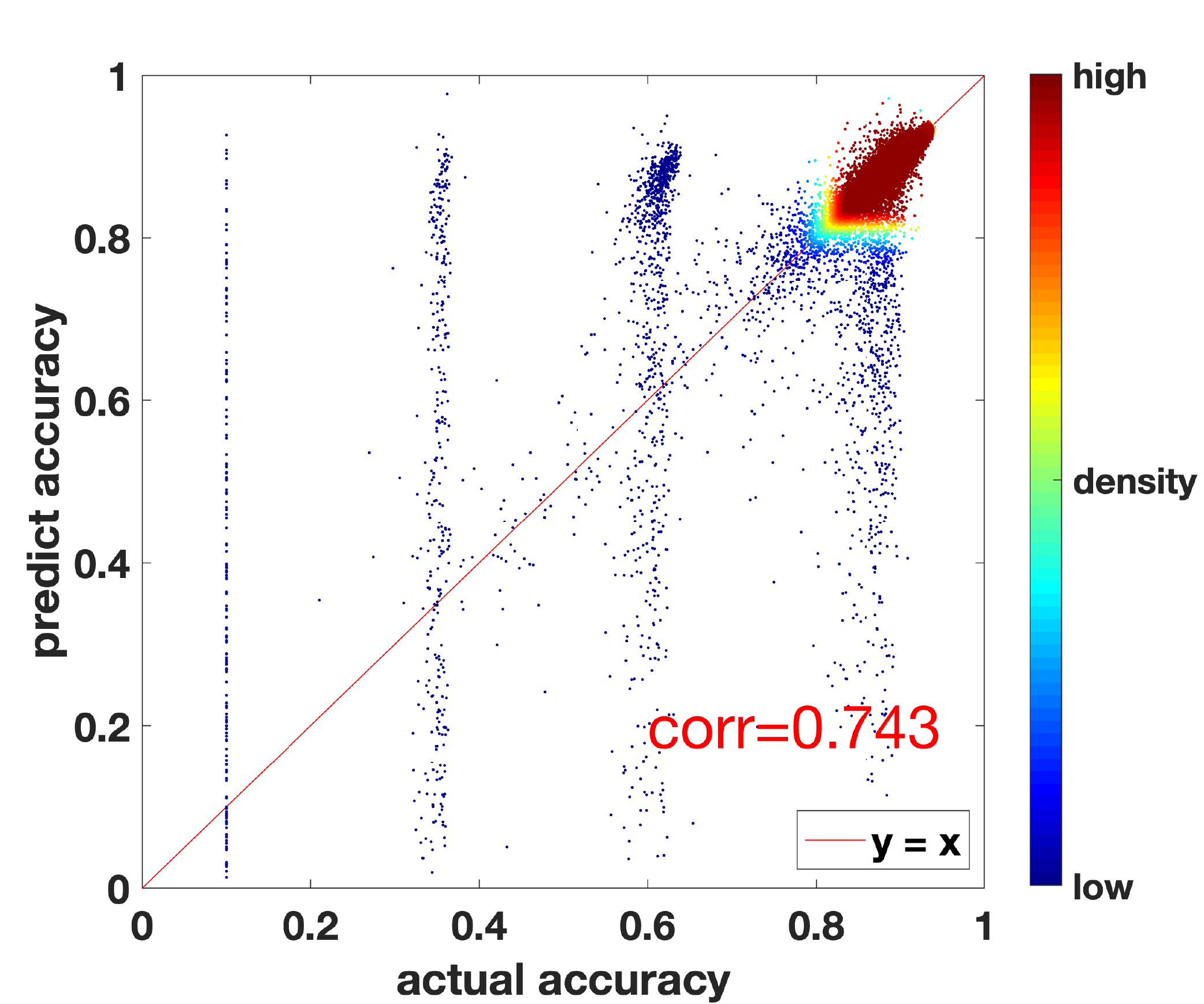}\label{fig:metadnn-rnn-test}} \\
  \subfloat[][MLP train]{ \includegraphics[width=0.49\columnwidth]{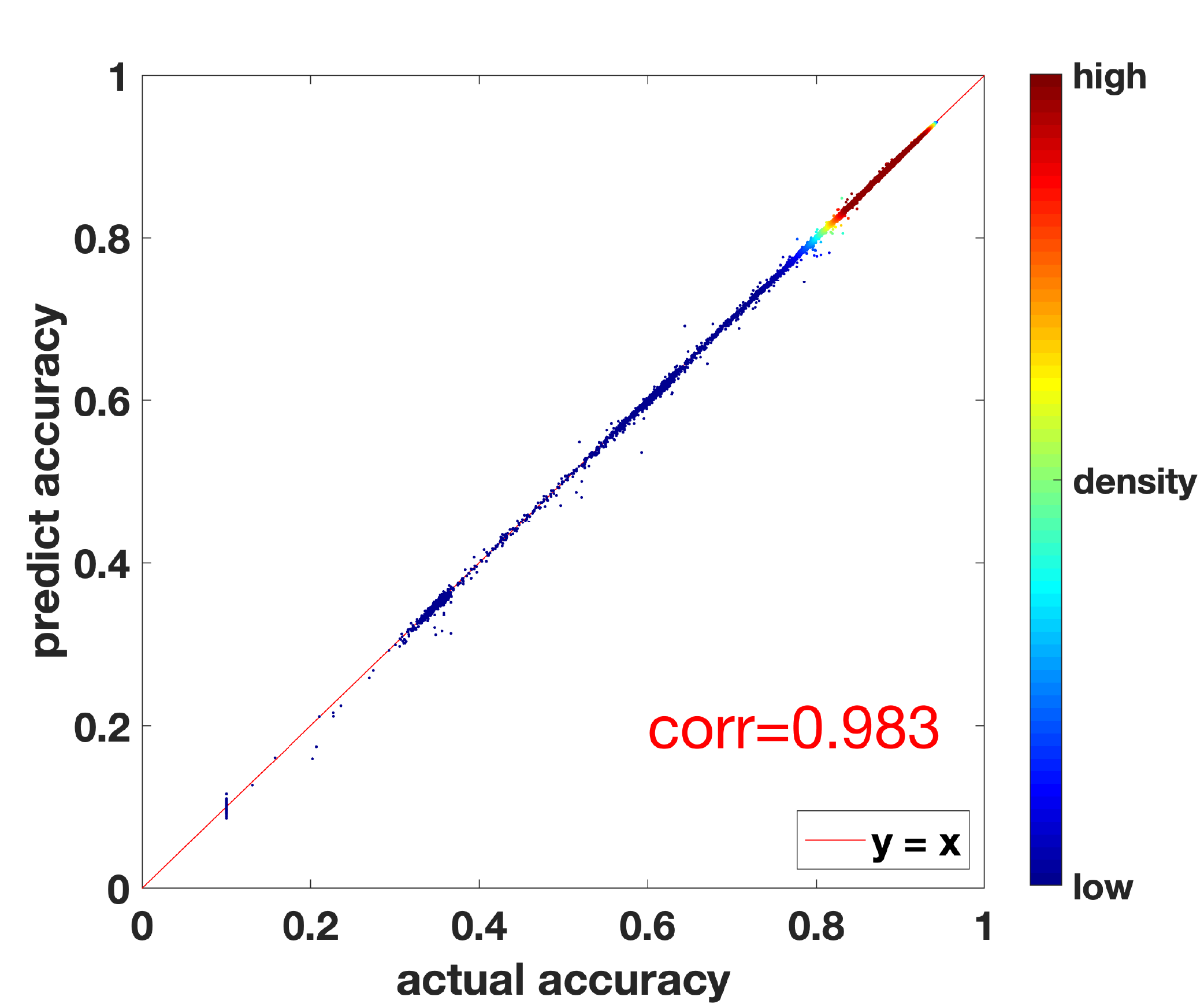}\label{fig:metadnn-mlp-train}} 
  \subfloat[][MLP test]{ \includegraphics[width=0.49\columnwidth]{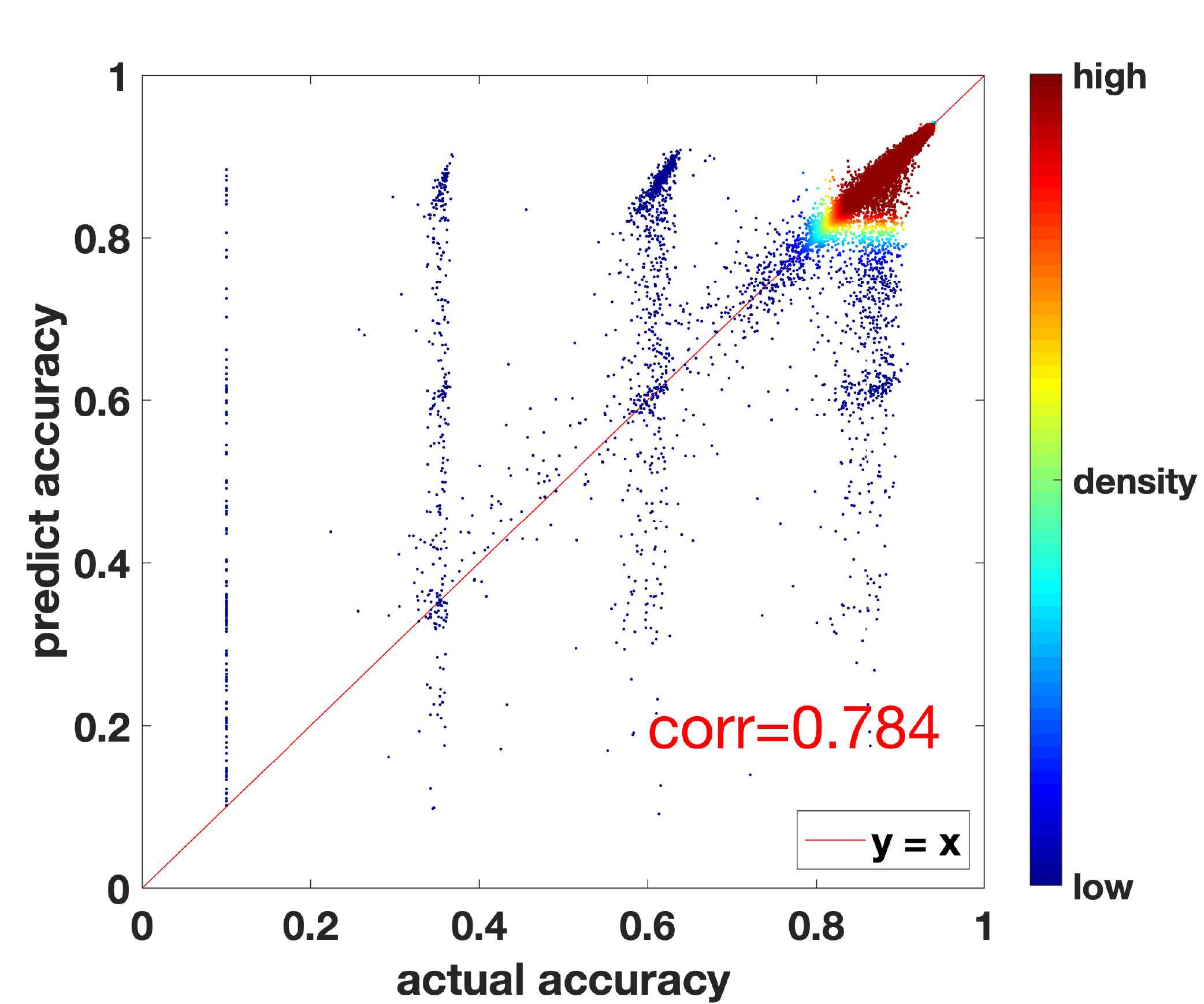}\label{fig:metadnn-mlp-test}} \\
  \subfloat[][multi-stage train]{ \includegraphics[width=0.49\columnwidth]{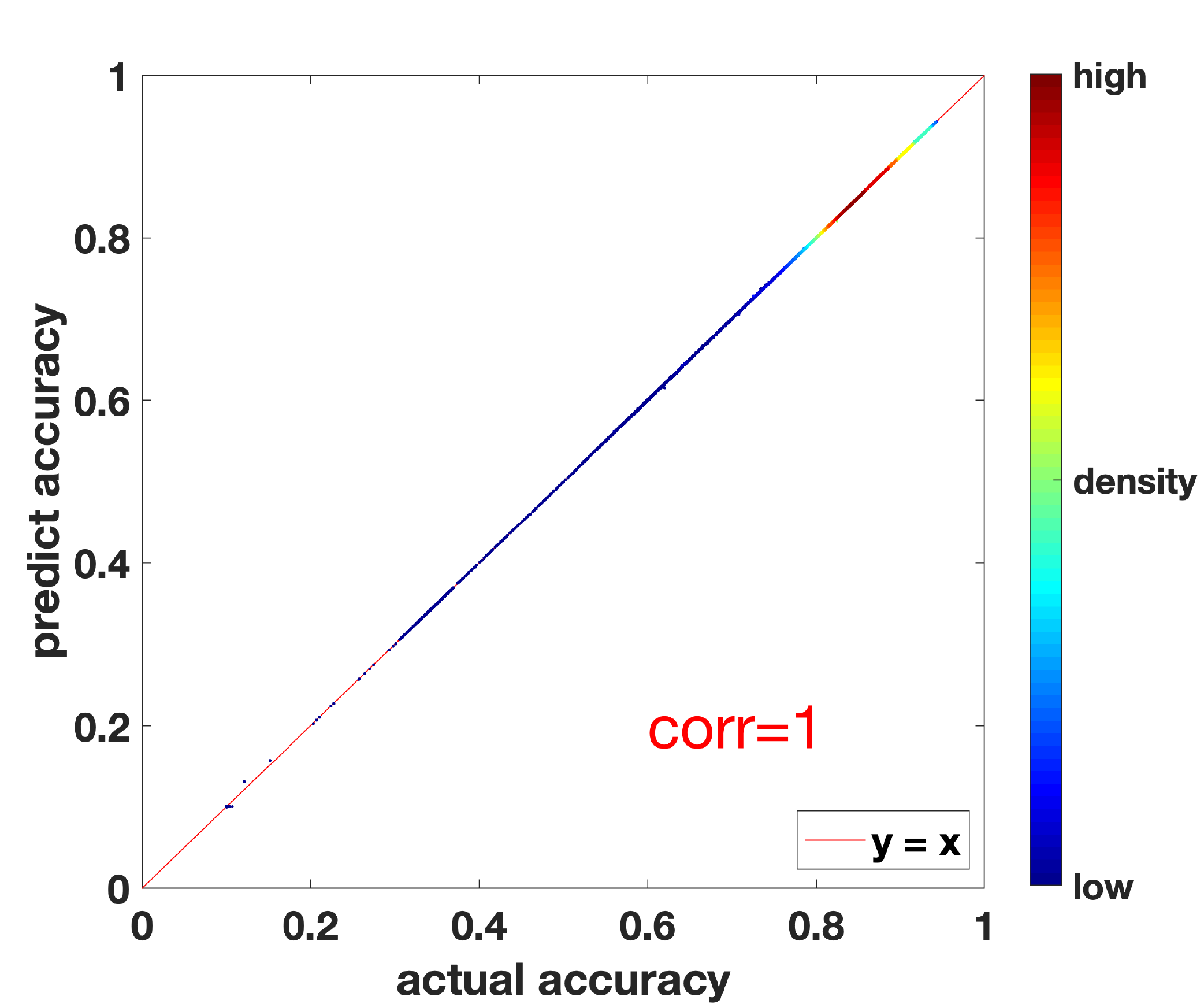}\label{fig:metadnn-mmlp-train}} 
  \subfloat[][multi-stage test]{ \includegraphics[width=0.49\columnwidth]{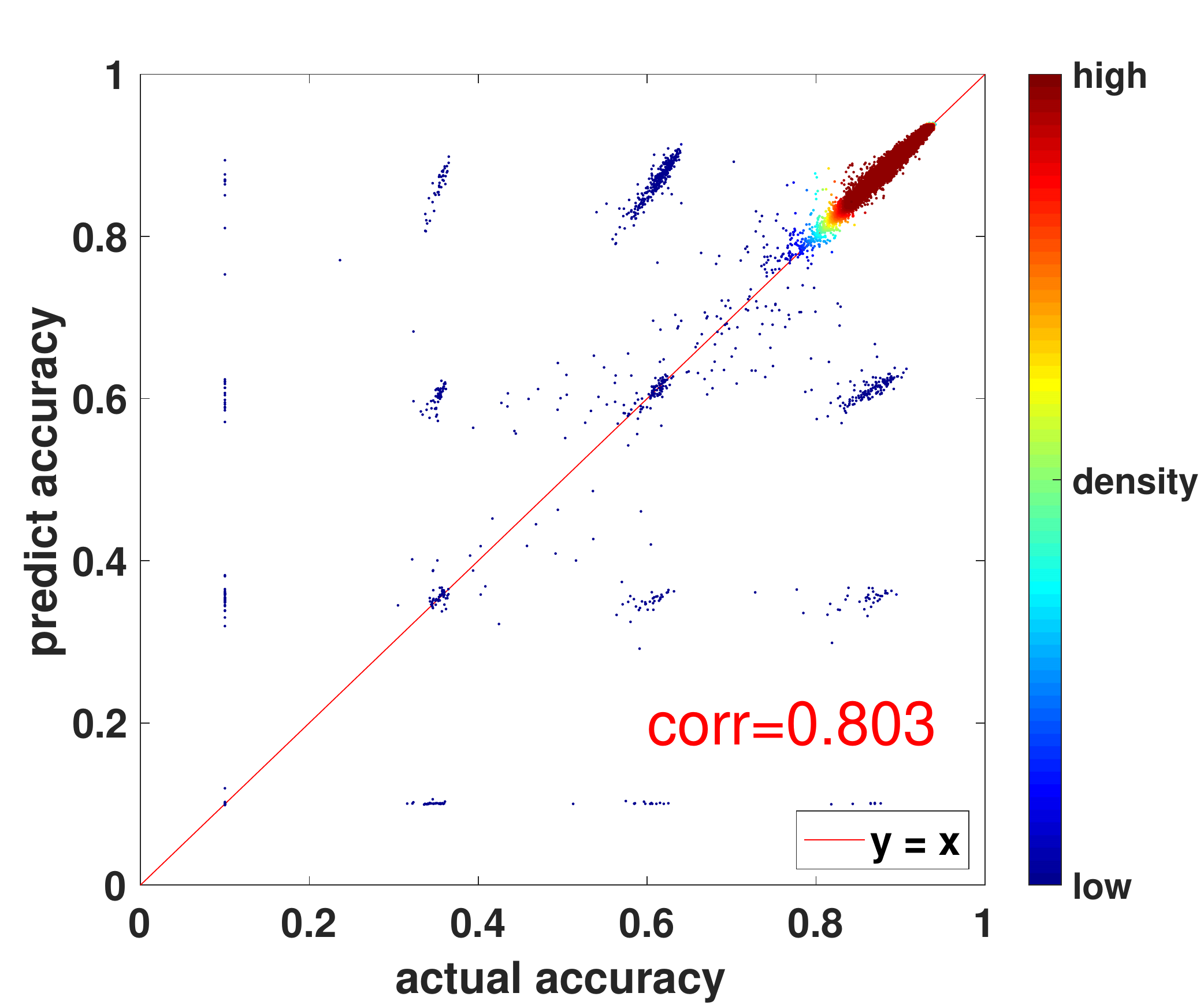}\label{fig:metadnn-mmlp-test}} 
  \caption{\textbf{meta-DNN design ablations}: True v.s. predicted accuracies of MLP, RNN and multi-stage MLP on architectures from NASBench. The scatter density is highlighted by color to reflect the data distribution; Red means high density, and blue otherwise. }
  \label{fig:metadnn-design}
  \end{figure}
  
\section{Experiments}

\subsection{Evaluations of architecture search}
\label{architecture_search_eval}

\textbf{\textit{Open domain search}}: we perform the search on CIFAR-10 using 8 NVIDIA 1080 TI. One GPU works as a server, while the rest work as clients. To further speedup network evaluations, we early terminated the training at 70th epoch during the pre-training. We selected the top 20 networks from the pre-training and fine-tuned them additional 530 epochs to get the final accuracy. For the ImageNet training, we constructed the network with the same $RCell$ and $NCell$ searched on CIFAR10 following the accepted standard, i.e. the mobile setting, defined in~\cite{zoph2017learning}. In total, AlphaX sampled 1000 networks; and Fig.~\ref{best-arch} demonstrates the architecture that yields the highest accuracy after fine-tuning. More details are available at appendix.~\ref{ap:experiment_setup}.

Table.~\ref{acc-comps-cifar} and Table.~\ref{acc-comps-imagenet} and summarize SOTA results on CIFAR10 and ImageNet, and AlphaX achieves SOTA accuracy with the least samples (M). Our end-to-end search cost, i.e. GPU days, is also on par with SOTA methods due to the early terminating and transfer learning. Notably, AlphaX achieves similar accuracy to AmoebaNet with 27x fewer samples for the case with cutout and filters = 32. Without cutout and filters = 32, AlphaX outperforms NAONet by 0.14\% in the test error with 16.7x fewer GPU days.

\textbf{\textit{Searching on NAS dataset}}: To further examine the sample efficiency, we evaluate AlphaX on the recent NAS dataset, NASBench-101~\cite{ying2019bench}. NASBench enumerates all the possible DAGs of nodes $\leq7$, constituting of (420k+) networks and their final test accuracies. This enables bypassing the computation barrier to fairly evaluate the sample efficiency. In our experiments, we limited the maximal nodes in a DAG $\leq$ 6, i.e. constructing a subset of NASBench-101 that contains 64521 valid networks. This allows us to quickly repeat each algorithm for 200 trials. The search target is the network with the highest mean test accuracy (the global optimum) at 108th epochs, which can be known ahead by querying the dataset. We choose Random Search (RS)~\cite{sciuto2019evaluating} and Regularized Evolution (RE)~\cite{real2018regularized} as the baseline, as RE delivers competitive results according to Table.~\ref{acc-comps-cifar} in the NASNet search space, and RS finds the global optimal in expected n/2, where n is the dataset size.

Fig.~\ref{fig:nasbench-results} demonstrates AlphaX is 2.8x and 3x faster than RE and RS, respectively. As we analyzed in Fig.~\ref{fig:teaser}, Random Search lacks an online model. Regularized Evolution only mutates on top-k performing models, while MCTS explicitly builds a search tree to dynamically trade off the exploration and exploitation at individual states. Please note that the slight difference in Fig.~\ref{fig:nasbench_best_trace} actually reflects a huge gap in speed as indicated by Fig.~\ref{fig:nasbench_samples_to_best}. From Fig.~\ref{fig:nasbench_acc_dist}, it shows there are abundant architectures with minor performance difference to the global optimum. Therefore, it is fast to find the top 5\% architectures, while slow in reaching the global optimum.

\textbf{\textit{Qualitative evaluations of AlphaX}}: Several interesting insights are observable in Fig.\ref{fig:mcts_viz}.
1) MCTS invests more on the promising directions (high-value path), and less otherwise (low-value path). Unlike greedy based algorithms, e.g. hill-climbing, MCTS consistently explores the search space guided by the adaptive UCT. 2) the best performing network is not necessarily located on the most promising branch, highlighting the importance of exploration in NAS.

\subsection{Component evaluations}
  
\textbf{\textit{Meta-DNN Design and its Impact}}:\label{sec:metadnn_design} The metric in evaluating metaDNN is the correlation between the predicted v.s. true accuracy. We used 80\% NASBench for training, and 20\% for testing. Since DNNs have shown great success in modeling complex data, we start with Multilayer Perceptron (MLP) and Recurrent Neural Network (RNN) on building the regression model. Specific architecture details are available in appendix.\ref{ap:metadnn-architecture}.
Fig.~\ref{fig:metadnn-rnn-test} and Fig~.\ref{fig:metadnn-mlp-test} demonstrate the performance of MLP (corr=0.784) is $4\%$ better than RNN (corr=0.743), as the MLP (Fig.~\ref{fig:metadnn-mlp-train}) performs much better than RNN (Fig.~\ref{fig:metadnn-rnn-train}) in the training set. However, MLP still mispredicts many networks around 0.1, 0.4 and 0.6 and 0.8 (x-axis) as shown in Fig.~\ref{fig:metadnn-mlp-test}. This clustering effect is consistent with the architecture distribution in Fig.~\ref{fig:nasbench_acc_dist} for having many networks around these accuracies. To alleviate this issue, we propose a multi-stage model, the core idea of which is to have several dedicated MLPs to predict different ranges of accuracies, e.g. [0, 25\%], along with another MLP to predict which MLP to use in predicting the final accuracy.
Fig.~\ref{fig:metadnn-mmlp-test} shows a multi-stage model successfully improves the correlation by 1.2\% from MLP, and the mispredictions have been greatly reduced. Since the multi-stage model has achieved corr = 1 on the training set, we choose it as the backbone regression model for AlphaX. Fig.~\ref{fig:nasbench-results} demonstrates our meta-DNN is effective to sustain NAS.

\textbf{\textit{Transfer Learning}}: the transfer learning significantly speeds network evaluations up, and Fig.~\ref{fig:transfer_learning} empirically validates the effectiveness of transfer learning. We randomly sampled an architecture as the parent network. On the parent network, we added a block with two new 5x5 separable conv layers on the left and right branch as the child network. We trained the parent network toward 70 epochs and saved its weights. In training the child network, we used weights from the parent network in initializing the child network except for two new conv layers that are randomly initialized. Fig.~\ref{fig:transfer_learning} shows the accuracy progress of transferred child network at different training epochs. The transferred child network retains the same accuracy as training from scratch (random initialization) with much fewer epochs, but insufficient epochs lose the accuracy. Therefore, we chose 20 epochs in pre-training an architecture if transfer learning applied.

\begin{figure}[t]
\begin{center}
  \subfloat[][best acc progression]{ \includegraphics[height=3.3cm]{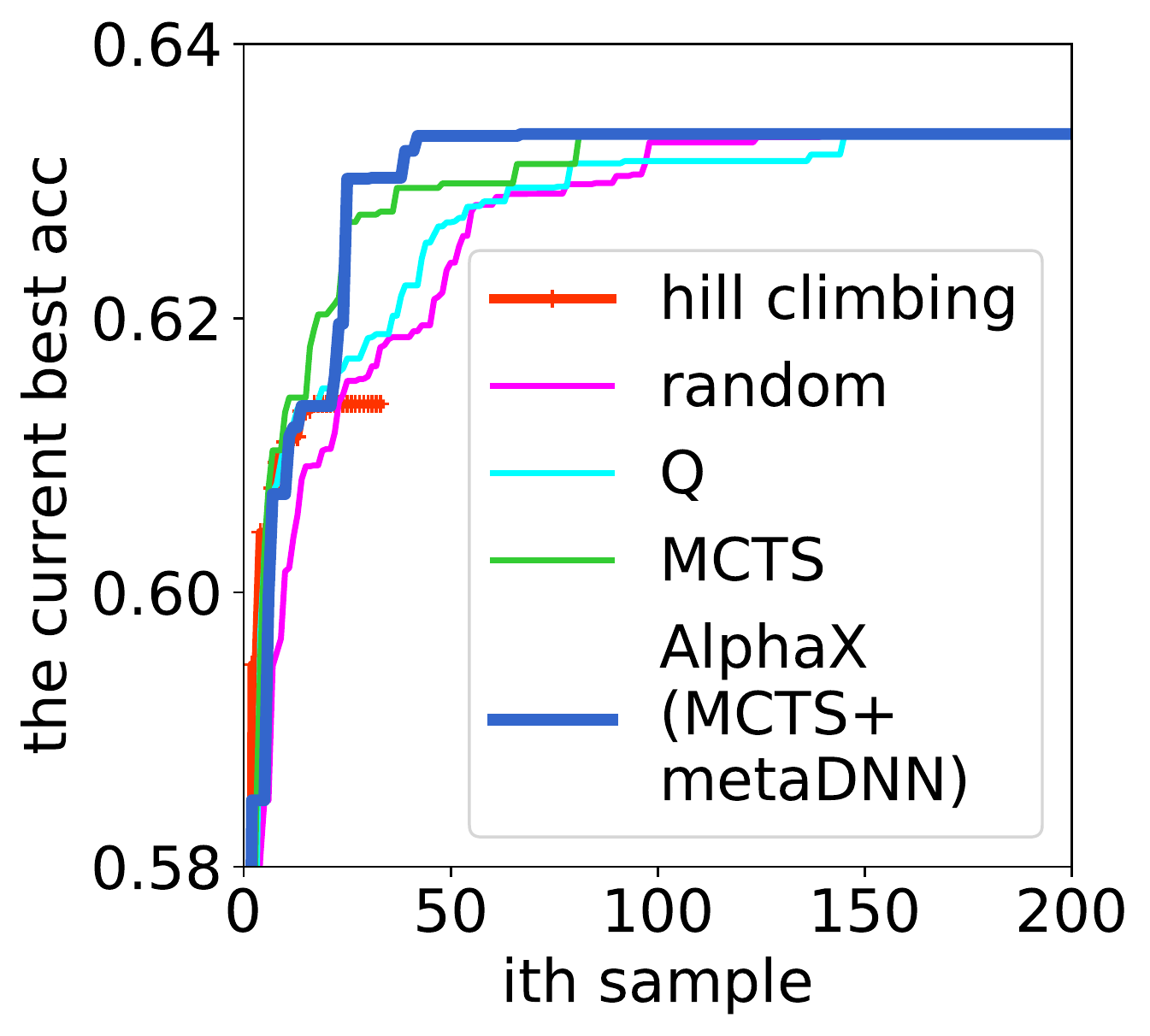}\label{fig:search-comparisions-line}}  \quad
\subfloat[][\#samples to the best]{ \includegraphics[height=3.2cm, trim={0.0cm 0cm 0cm 0.0cm}]{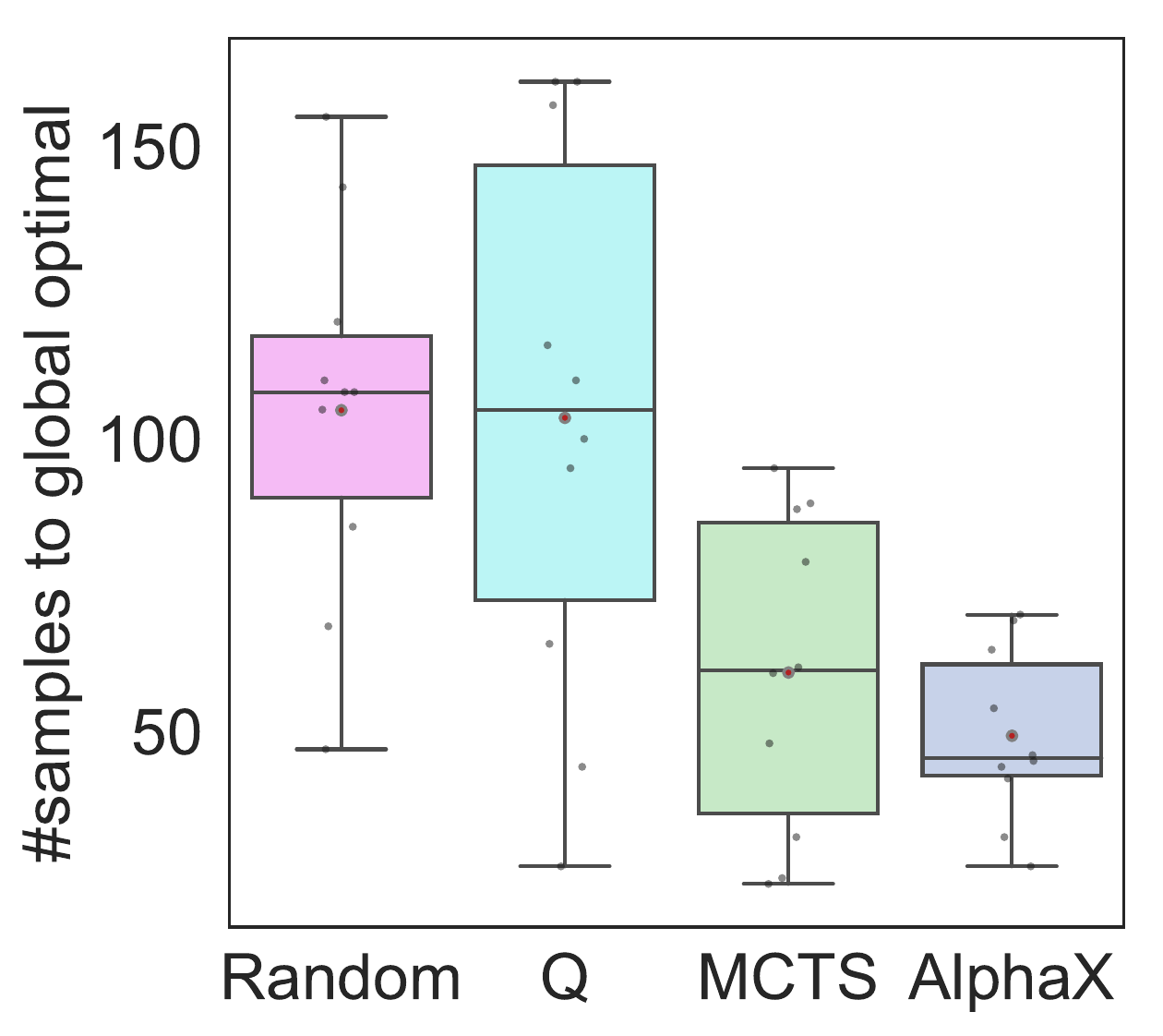}\label{fig:search-comparisions-boxplot}}
\end{center}
\caption{\textbf{Algorithmic comparisons}: AlphaX is consistently the fastest algorithm to reach the global optimal on another simplified search domain (appendix.\ref{ap:comparison}), while Hill Climbing can easily trap into a local optimal.  }
\label{fig:alg_comps}
\end{figure}

\subsection{Algorithm Comparisons}
\label{alg-comps-sec}

Fig.~\ref{fig:alg_comps} evaluates MCTS against Q-Learning (QL), Hill Climbing (HC) and Random Search (RS) on a simplified search space. Setup details and the introduction of the design domain are available in appendix.\ref{ap:comparison}. These algorithms are widely used in NAS \cite{Baker2016,Liu2017,elsken2017simple,greff2017lstm}. We conduct 10 trials for each algorithm. Fig.~\ref{fig:search-comparisions-boxplot} demonstrates AlphaX is 2.3x faster than QL and RS. Though HC is the fastest, Fig.~\ref{fig:search-comparisions-line} indicates HC traps into a local optimal. Interestingly, Fig.~\ref{fig:search-comparisions-boxplot} indicates the inter-quartile range of QL is longer than RS. This is because QL quickly converges to a suboptimal, spending a huge time to escape. This is consistent with Fig.~\ref{fig:search-comparisions-line} that QL converges faster than RS before the 50th samples, but random can easily escape from the local optimal afterward. Fig.~\ref{fig:search-comparisions-boxplot} (MCTS v.s. AlphaX) further corroborates the effectiveness of meta-DNN.

\begin{figure}[t]
\centering
\subfloat[][content]{\includegraphics[width=0.4\columnwidth]{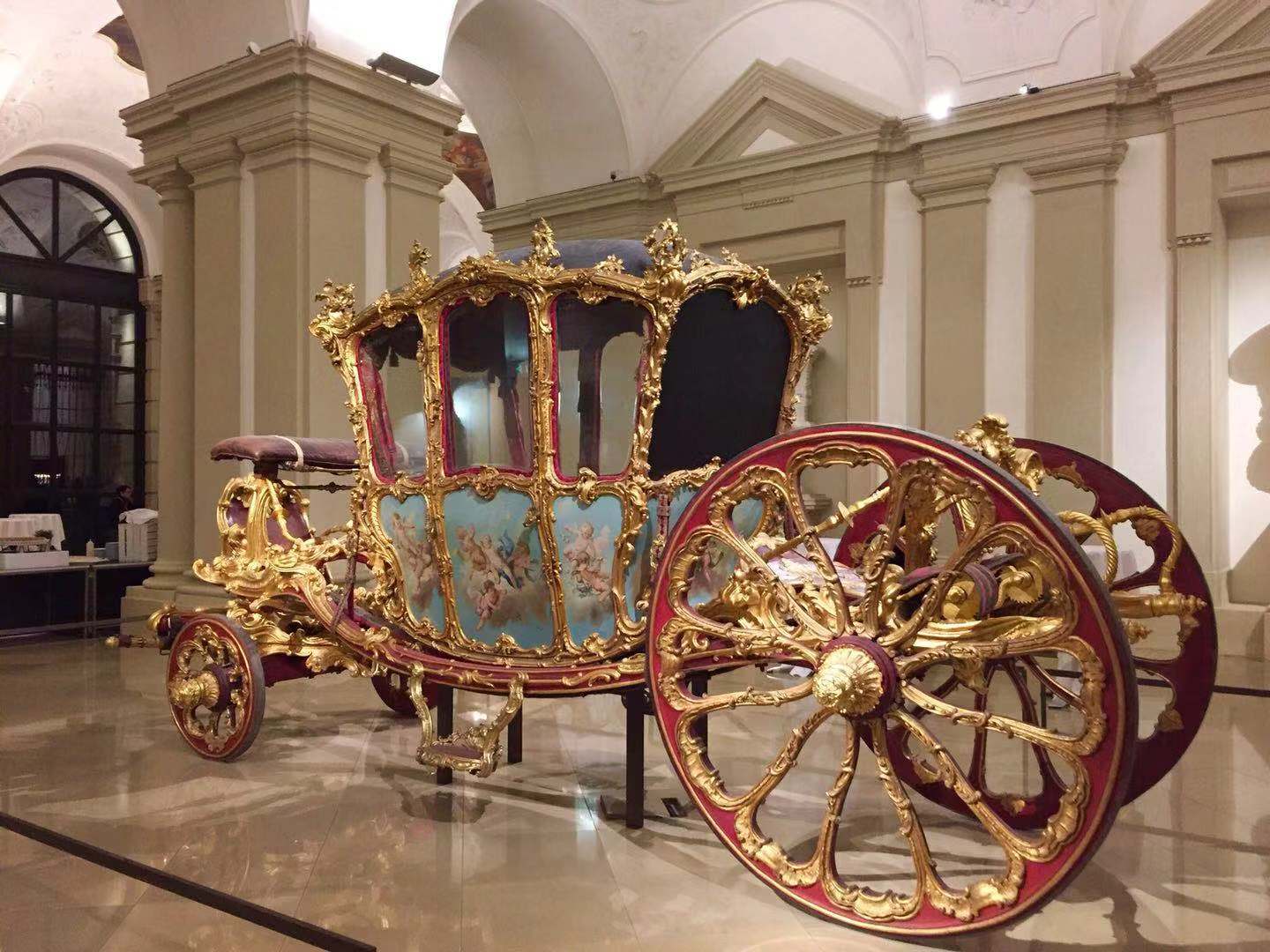}} \quad
\subfloat[][style]{\includegraphics[width=0.4\columnwidth]{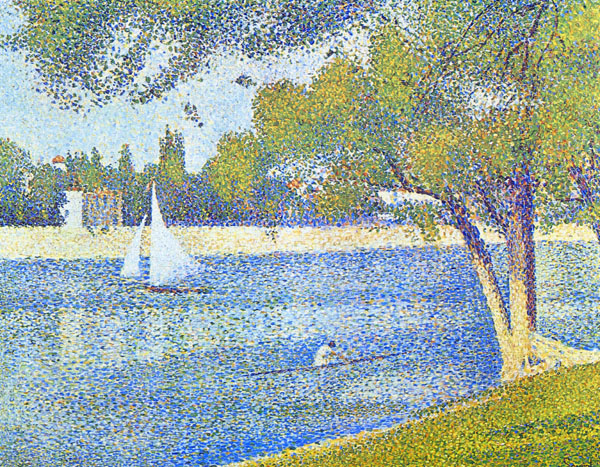}} \\
\subfloat[][VGG]{\includegraphics[width=0.4\columnwidth]{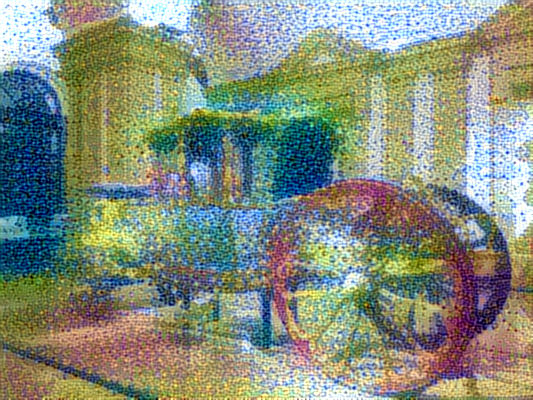}} \quad
\subfloat[][AlphaX-1]{\includegraphics[width=0.4\columnwidth]{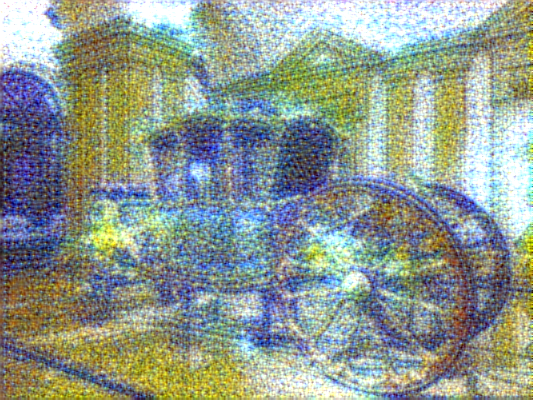}}
\caption{ \textbf{Neural Style Transfer}: AlphaX-1 v.s. VGG.}
\label{neural-style-transfer}
\end{figure}

\subsection{Improved Features for Vision Applications}

CNN is a common component for Computer Vision (CV) models. Here, we demonstrate the searched architecture can improve a variety of downstream Computer Vision (CV) applications. Please check the Appendix Sec.\ref{ap:setup_vision_models} for the experiment setup. 

1) \textit{Neural Style Transfer}: AlphaX-1 is better than a shallow network (VGG) in capturing the rich details and textures of a sophisticated style image (Fig.~\ref{neural-style-transfer}). 

\begin{figure}[h]
  \centering
  \subfloat[][AlphaX-1]{\includegraphics[height=1.3in, trim={6cm 8cm 0cm 5cm},clip]{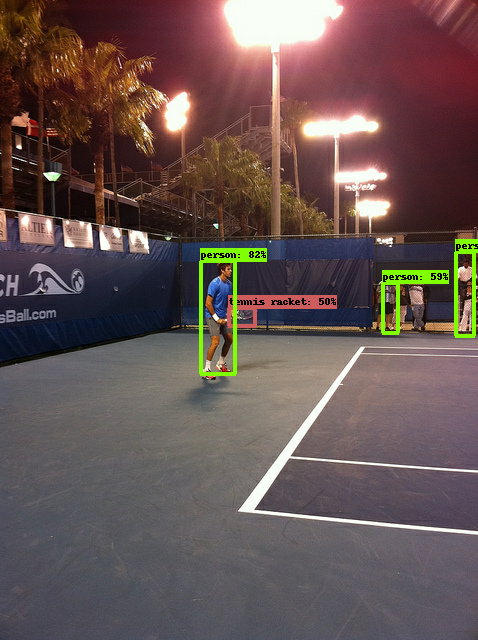}\label{MCTS_state}} \quad
  \subfloat[][MobileNet-v1]{\includegraphics[height=1.3in, trim={6cm 8cm 0cm 5cm},clip]{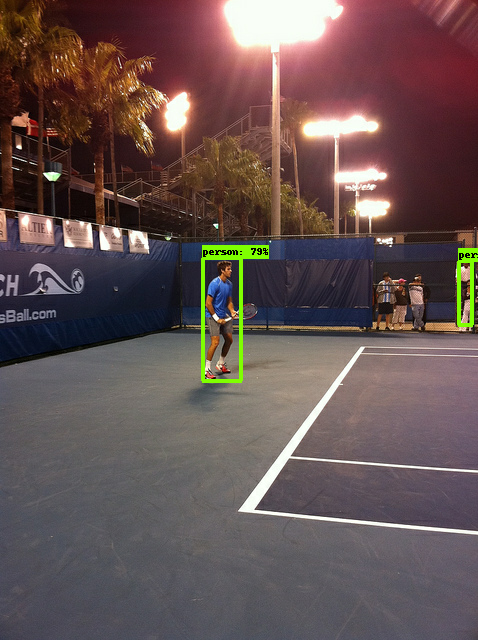}\label{MCTS_state}} \quad \\
  \subfloat[][AlphaX-1]{\includegraphics[height=1.3in, trim={6cm 3cm 6cm 4cm},clip]{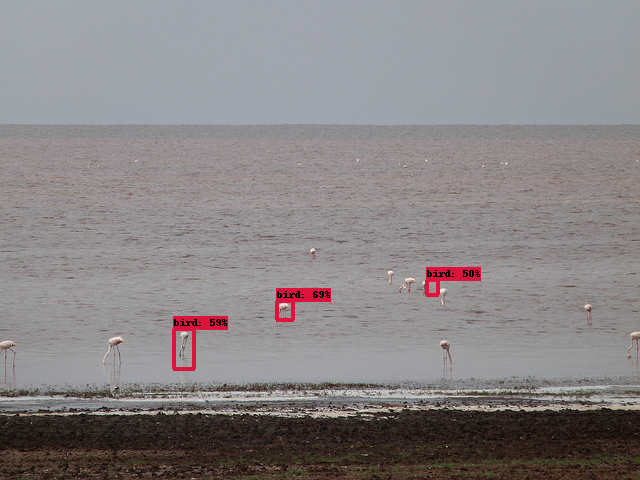}\label{MCTS_state}} \quad
  \subfloat[][MobileNet-v1]{\includegraphics[height=1.3in, trim={6cm 3cm 6cm 4cm},clip]{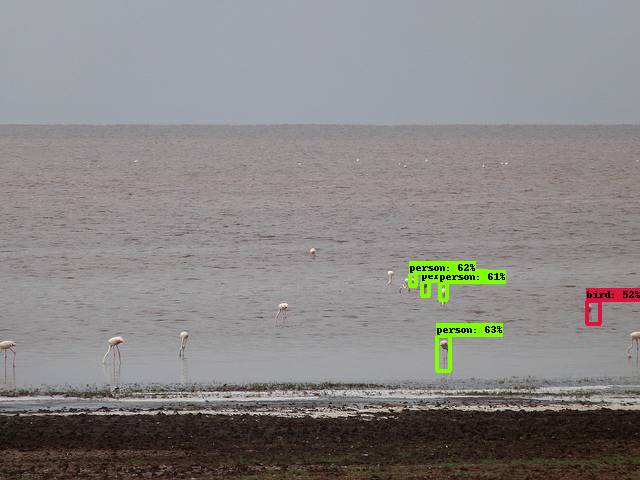}\label{MCTS_state}} \quad
  \caption{\textit{Object Detection}: the object detection system is more precise with AlphaX-1 than MobileNet. }
  \label{object-detection}
\end{figure}

2) \textit{Object Detection}: We replace MobileNet-v1 with AlphaX-1 in SSD~\cite{liu2016ssd} object detection model, and the mAP (mini-val) increases from $20.1\%$ to $23.7\%$ at the $300\times300$ resolution. (Fig.\ref{object-detection}, appendix). 

\begin{figure}[t]
  \subfloat[][VGG: a tennis player is playing tennis on the court. \\
        AlphaX-1: a couple of people playing tennis on a tennis court.        ]{\includegraphics[height=1in]{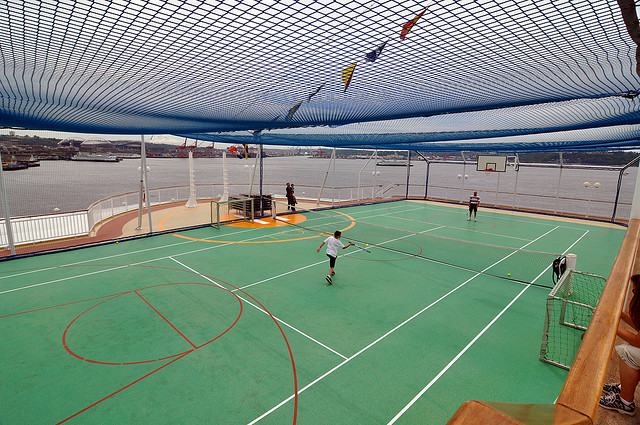}\label{MCTS_state}} \quad
  \subfloat[][VGG: a bus is parked on the side of the road. \\ AlphaX-1: a blue bus is driving down a street.]{\includegraphics[height=1in]{COCO_val2014_000000133819.jpg}\label{MCTS_state}} \quad
  \subfloat[][VGG: a cup of coffee and a cup of coffee. \\
  AlphaX-1: a book and a cup of coffee on a table. ]{\includegraphics[height=1in]{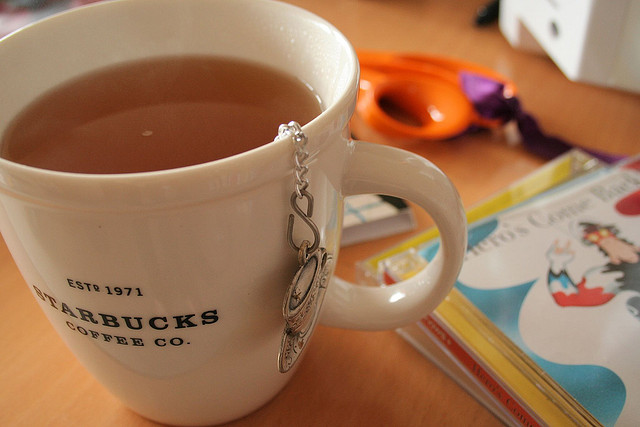}\label{MCTS_state}} \quad
  \subfloat[][VGG: a fire hydrant on the side of a road.\\
  AlphaX-1: a red fire hydrant on the side of a road.
  ]{\includegraphics[height=1in]{COCO_val2014_000000398097.jpg}\label{MCTS_state}}\quad
  \caption{ \textit{Image Captioning}: AlphaX-1 captures more details than VGG in the captions.}
  \label{fig:captions}
\end{figure}

3) \textit{Image Captioning}: we replace the VGG with AlphaX-1 in \textit{show attend and tell}~\cite{xu2015show}. On the 2014 MSCOCO-val dataset, AlphaX-1 outperforms VGG by 2.4 (RELU-2), 4.4 (RELU-3), 3.7 (RELU-4), respectively (Fig.\ref{fig:captions}, appendix).

\section{Conclusion}
In this paper, we propose a new MCTS based NAS agent named AlphaX. Compared to prior MCTS agents, AlphaX is the first practical MCTS based agent that achieves SOTA results on both CIFAR-10 and ImageNet in a reasonable amount of computations by improving the sampling efficiency with a novel predictive model meta-DNN, and by amortizing the network evaluation costs with a scalable solution and transfer learning. In 3 search tasks, AlphaX consistently demonstrates superior search efficiency over mainstream algorithms, highlighting MCTS as a promising search algorithm for NAS.


\bibliographystyle{aaai}
\bibliography{egbib}

\begin{thebibliography}{}

\bibitem[\protect\citeauthoryear{Auer, Cesa-Bianchi, and
  Fischer}{2002}]{auer2002finite}
Auer, P.; Cesa-Bianchi, N.; and Fischer, P.
\newblock 2002.
\newblock Finite-time analysis of the multiarmed bandit problem.
\newblock {\em Machine learning} 47(2-3):235--256.

\bibitem[\protect\citeauthoryear{Baker \bgroup et al\mbox.\egroup
  }{2016}]{Baker2016}
Baker, B.; Gupta, O.; Naik, N.; and Raskar, R.
\newblock 2016.
\newblock {Designing Neural Network Architectures using Reinforcement
  Learning}.
\newblock  1--18.

\bibitem[\protect\citeauthoryear{Bergstra \bgroup et al\mbox.\egroup
  }{2011}]{bergstra2011algorithms}
Bergstra, J.~S.; Bardenet, R.; Bengio, Y.; and K{\'e}gl, B.
\newblock 2011.
\newblock Algorithms for hyper-parameter optimization.
\newblock In {\em Advances in neural information processing systems},
  2546--2554.

\bibitem[\protect\citeauthoryear{Chen \bgroup et al\mbox.\egroup
  }{2019}]{chen2019renas}
Chen, Y.; Meng, G.; Zhang, Q.; Xiang, S.; Huang, C.; Mu, L.; and Wang, X.
\newblock 2019.
\newblock Renas: Reinforced evolutionary neural architecture search.
\newblock In {\em Proceedings of the IEEE Conference on Computer Vision and
  Pattern Recognition},  4787--4796.

\bibitem[\protect\citeauthoryear{Elsken, Metzen, and
  Hutter}{2017}]{elsken2017simple}
Elsken, T.; Metzen, J.-H.; and Hutter, F.
\newblock 2017.
\newblock Simple and efficient architecture search for convolutional neural
  networks.
\newblock {\em arXiv preprint arXiv:1711.04528}.

\bibitem[\protect\citeauthoryear{Gatys, Ecker, and Bethge}{2015}]{neural_style}
Gatys, L.~A.; Ecker, A.~S.; and Bethge, M.
\newblock 2015.
\newblock A neural algorithm of artistic style.
\newblock {\em CoRR} abs/1508.06576.

\bibitem[\protect\citeauthoryear{Greff \bgroup et al\mbox.\egroup
  }{2017}]{greff2017lstm}
Greff, K.; Srivastava, R.~K.; Koutn{\'\i}k, J.; Steunebrink, B.~R.; and
  Schmidhuber, J.
\newblock 2017.
\newblock Lstm: A search space odyssey.
\newblock {\em IEEE transactions on neural networks and learning systems}
  28(10):2222--2232.

\bibitem[\protect\citeauthoryear{Kandasamy, Schneider, and
  P{\'o}czos}{2015}]{kandasamy2015high}
Kandasamy, K.; Schneider, J.; and P{\'o}czos, B.
\newblock 2015.
\newblock High dimensional bayesian optimisation and bandits via additive
  models.
\newblock In {\em International Conference on Machine Learning},  295--304.

\bibitem[\protect\citeauthoryear{Kocsis and
  Szepesv{\'a}ri}{2006}]{kocsis2006bandit}
Kocsis, L., and Szepesv{\'a}ri, C.
\newblock 2006.
\newblock Bandit based monte-carlo planning.
\newblock In {\em European conference on machine learning},  282--293.
\newblock Springer.

\bibitem[\protect\citeauthoryear{Lin \bgroup et al\mbox.\egroup
  }{2014a}]{LinMBHPRDZ14}
Lin, T.; Maire, M.; Belongie, S.~J.; Bourdev, L.~D.; Girshick, R.~B.; Hays, J.;
  Perona, P.; Ramanan, D.; Doll{\'{a}}r, P.; and Zitnick, C.~L.
\newblock 2014a.
\newblock Microsoft {COCO:} common objects in context.
\newblock {\em CoRR} abs/1405.0312.

\bibitem[\protect\citeauthoryear{Lin \bgroup et al\mbox.\egroup
  }{2014b}]{MSCOCO2014}
Lin, T.; Maire, M.; Belongie, S.~J.; Bourdev, L.~D.; Girshick, R.~B.; Hays, J.;
  Perona, P.; Ramanan, D.; Doll{\'{a}}r, P.; and Zitnick, C.~L.
\newblock 2014b.
\newblock Microsoft {COCO:} common objects in context.
\newblock {\em CoRR} abs/1405.0312.

\bibitem[\protect\citeauthoryear{Liu \bgroup et al\mbox.\egroup
  }{2016}]{liu2016ssd}
Liu, W.; Anguelov, D.; Erhan, D.; Szegedy, C.; Reed, S.; Fu, C.-Y.; and Berg,
  A.~C.
\newblock 2016.
\newblock Ssd: Single shot multibox detector.
\newblock In {\em European conference on computer vision},  21--37.
\newblock Springer.

\bibitem[\protect\citeauthoryear{Liu \bgroup et al\mbox.\egroup
  }{2017a}]{liu2017progressive}
Liu, C.; Zoph, B.; Shlens, J.; Hua, W.; Li, L.-J.; Fei-Fei, L.; Yuille, A.;
  Huang, J.; and Murphy, K.
\newblock 2017a.
\newblock Progressive neural architecture search.
\newblock {\em arXiv preprint arXiv:1712.00559}.

\bibitem[\protect\citeauthoryear{Liu \bgroup et al\mbox.\egroup
  }{2017b}]{Liu2017}
Liu, C.; Zoph, B.; Shlens, J.; Hua, W.; Li, L.-J.; Fei-Fei, L.; Yuille, A.;
  Huang, J.; and Murphy, K.
\newblock 2017b.
\newblock {Progressive Neural Architecture Search}.

\bibitem[\protect\citeauthoryear{Liu \bgroup et al\mbox.\egroup
  }{2017c}]{liu2017hierarchical}
Liu, H.; Simonyan, K.; Vinyals, O.; Fernando, C.; and Kavukcuoglu, K.
\newblock 2017c.
\newblock Hierarchical representations for efficient architecture search.
\newblock {\em arXiv preprint arXiv:1711.00436}.

\bibitem[\protect\citeauthoryear{Liu \bgroup et al\mbox.\egroup
  }{2018}]{liu2018darts}
Liu, H.; Simonyan, K.; Simonyan, K.; and Yang, Y.
\newblock 2018.
\newblock Darts: Differentiable architecture search.
\newblock {\em arXiv preprint arXiv:1806.09055}.

\bibitem[\protect\citeauthoryear{Loshchilov and Hutter}{2017}]{cosine_restart}
Loshchilov, I., and Hutter, F.
\newblock 2017.
\newblock Sgdr: Stochastic gradient descent with warm restarts.
\newblock {\em arXiv preprint arXiv:1608.03983}.

\bibitem[\protect\citeauthoryear{Luo \bgroup et al\mbox.\egroup
  }{2018}]{luo2018neural}
Luo, R.; Tian, F.; Qin, T.; Chen, E.; and Liu, T.-Y.
\newblock 2018.
\newblock Neural architecture optimization.
\newblock In Bengio, S.; Wallach, H.; Larochelle, H.; Grauman, K.;
  Cesa-Bianchi, N.; and Garnett, R., eds., {\em Advances in Neural Information
  Processing Systems 31}. Curran Associates, Inc.
\newblock  7816--7827.

\bibitem[\protect\citeauthoryear{Nachum, Norouzi, and
  Schuurmans}{2016}]{nachum2016improving}
Nachum, O.; Norouzi, M.; and Schuurmans, D.
\newblock 2016.
\newblock Improving policy gradient by exploring under-appreciated rewards.
\newblock {\em arXiv preprint arXiv:1611.09321}.

\bibitem[\protect\citeauthoryear{Negrinho and
  Gordon}{2017}]{negrinho2017deeparchitect}
Negrinho, R., and Gordon, G.
\newblock 2017.
\newblock Deeparchitect: Automatically designing and training deep
  architectures.
\newblock {\em arXiv preprint arXiv:1704.08792}.

\bibitem[\protect\citeauthoryear{Pham \bgroup et al\mbox.\egroup
  }{2018}]{pham2018efficient}
Pham, H.; Guan, M.~Y.; Zoph, B.; Le, Q.~V.; and Dean, J.
\newblock 2018.
\newblock Efficient neural architecture search via parameter sharing.
\newblock {\em arXiv preprint arXiv:1802.03268}.

\bibitem[\protect\citeauthoryear{Real \bgroup et al\mbox.\egroup
  }{2018}]{real2018regularized}
Real, E.; Aggarwal, A.; Huang, Y.; and Le, Q.~V.
\newblock 2018.
\newblock Regularized evolution for image classifier architecture search.
\newblock {\em arXiv preprint arXiv:1802.01548}.

\bibitem[\protect\citeauthoryear{Sciuto \bgroup et al\mbox.\egroup
  }{2019}]{sciuto2019evaluating}
Sciuto, C.; Yu, K.; Jaggi, M.; Musat, C.; and Salzmann, M.
\newblock 2019.
\newblock Evaluating the search phase of neural architecture search.
\newblock {\em arXiv preprint arXiv:1902.08142}.

\bibitem[\protect\citeauthoryear{Silver \bgroup et al\mbox.\egroup
  }{2016}]{silver2016mastering}
Silver, D.; Huang, A.; Maddison, C.~J.; Guez, A.; Sifre, L.; Van Den~Driessche,
  G.; Schrittwieser, J.; Antonoglou, I.; Panneershelvam, V.; Lanctot, M.;
  et~al.
\newblock 2016.
\newblock Mastering the game of go with deep neural networks and tree search.
\newblock {\em nature} 529(7587):484--489.

\bibitem[\protect\citeauthoryear{Szegedy \bgroup et al\mbox.\egroup
  }{2015}]{SzegedyVISW15}
Szegedy, C.; Vanhoucke, V.; Ioffe, S.; Shlens, J.; and Wojna, Z.
\newblock 2015.
\newblock Rethinking the inception architecture for computer vision.
\newblock {\em CoRR} abs/1512.00567.

\bibitem[\protect\citeauthoryear{Tian \bgroup et al\mbox.\egroup
  }{2019}]{yuandong2019OpeGo}
Tian, Y.; Ma, J.; Gong, Q.; Sengupta, S.; Chen, Z.; Pinkerton, J.; and Zitnick,
  C.~L.
\newblock 2019.
\newblock {ELF} opengo: An analysis and open reimplementation of alphazero.
\newblock {\em CoRR} abs/1902.04522.

\bibitem[\protect\citeauthoryear{Wistuba}{2017}]{wistuba2017finding}
Wistuba, M.
\newblock 2017.
\newblock Finding competitive network architectures within a day using uct.
\newblock {\em arXiv preprint arXiv:1712.07420}.

\bibitem[\protect\citeauthoryear{Xu \bgroup et al\mbox.\egroup
  }{2015}]{xu2015show}
Xu, K.; Ba, J.; Kiros, R.; Cho, K.; Courville, A.; Salakhudinov, R.; Zemel, R.;
  and Bengio, Y.
\newblock 2015.
\newblock Show, attend and tell: Neural image caption generation with visual
  attention.
\newblock In {\em International conference on machine learning},  2048--2057.

\bibitem[\protect\citeauthoryear{Ying \bgroup et al\mbox.\egroup
  }{2019}]{ying2019bench}
Ying, C.; Klein, A.; Real, E.; Christiansen, E.; Murphy, K.; and Hutter, F.
\newblock 2019.
\newblock Nas-bench-101: Towards reproducible neural architecture search.
\newblock {\em arXiv preprint arXiv:1902.09635}.

\bibitem[\protect\citeauthoryear{Zoph and Le}{2016}]{Zoph2016}
Zoph, B., and Le, Q.~V.
\newblock 2016.
\newblock {Neural Architecture Search with Reinforcement Learning}.
\newblock  1--16.

\bibitem[\protect\citeauthoryear{Zoph \bgroup et al\mbox.\egroup
  }{2017}]{zoph2017learning}
Zoph, B.; Vasudevan, V.; Shlens, J.; and Le, Q.~V.
\newblock 2017.
\newblock Learning transferable architectures for scalable image recognition.
\newblock {\em arXiv preprint arXiv:1707.07012}.

\end{thebibliography}

\newpage

\renewcommand\thesection{\Alph{section}}
\setcounter{section}{0}
\section{Pseudocode of AlphaX}
\label{ap:pseudocode}

In this section, we present the pseudocode of distributed AlphaX. Algorithm~\ref{MCTS} describes the search engine;
Algorithm~\ref{server} describes the procedures of server that implement a client-server communication protocol to send an architecture searched by MCTS to a client for training.   
Algorithm~\ref{client} describes the procedures of client that consistently listen new architectures from the server for training, and send the accuracy and network back to the server after training.

\begin{algorithm}[H]
  \caption{Client}
  \label{client}
    \begin{algorithmic}[1]
      \STATE {\bfseries Require:} Start working once building connection to the server
      \WHILE{True}
      \IF{The client is connected to server}
       \STATE $network \leftarrow \text{Receive}()$
       \STATE $accuracy \leftarrow \text{Train}(network)$
       \STATE Send $(network, accuracy)$ to the Server
      \ELSE
      \STATE Wait for re-connection
      \ENDIF
      \ENDWHILE
    \end{algorithmic}
\end{algorithm}

\begin{algorithm}[H]
  \caption{Server}
  \label{server}
    \begin{algorithmic}[1]       
      \WHILE{$size(TASK_{-}QUEUE) > 2$}
      \WHILE{no idle client}
      \STATE Continue \Commentl{Wait for dispatching jobs until there are idle clients}
      \ENDWHILE
      \STATE Create a new connection to a random idle client
      \STATE $network \leftarrow TASK_{-}QUEUE.\text{pop}()$
      \STATE Send $network$ to a Client
      \IF{$\text{Received}_{-}\text{Signal}()$}
      \STATE $network, accuracy \leftarrow \text{Receive}_{-}\text{Result}()$
      \STATE $acc(network) \leftarrow accuracy$
      \STATE $state \leftarrow rollout_{-}from(network)$
      \STATE Backpropagation($state$, $(accuracy - \hat{q}(state)) / 2$, 0) \Comment{Replace $\hat{q}$ with $q=Q(s,a)$ in Eq. \ref{reward}}
      \STATE Train the meta-DNN with a new data $(network, accuracy)$
      \ELSE 
      \STATE Continue 
      \ENDIF
      \ENDWHILE
    \end{algorithmic}
\end{algorithm}

\newpage

\begin{algorithm}[H]
  \caption{Search Engine (MCTS)}
  \label{MCTS}
    \begin{algorithmic}[1]
    \STATE \textbf{function} {Expansion}($state$)
    \STATE \quad Create a new node in a tree for state.
    \STATE \quad \textbf{for all} $action$ available at $state$ \textbf{do}
    \STATE \quad $Q(state, action) \leftarrow 0,\quad N(state, action) \leftarrow 0$ 
    \STATE \quad \textbf{end for}
    \STATE \textbf{end function}
    \STATE
    \STATE \textbf{function} {Simulation}($state$)
    \STATE $action \leftarrow none$
    \STATE \quad \textbf{while} $action\;is\;not\;term$ \textbf{do}
    \STATE \quad \quad randomly generate an $action$
    \STATE \quad \quad $next_{-}net$ $\leftarrow$ $\text{Apply}(state, action)$ \Comment{$\text{Apply}$ returns the next state when $action$ is applied to $state$}
    \STATE \quad \textbf{end while}
    \STATE \textbf{end function}
    \STATE
    \STATE \textbf{function} {Backpropagation}($state$, $q$, $n$)
    \STATE \quad \textbf{while} $state\;is\;not\;root$ \textbf{do}
    \STATE \quad \quad $state$ $\leftarrow$ $\text{parent}(state)$
    \STATE \quad \quad $Q(state, action) \leftarrow Q(state, action) + q$
    \STATE \quad \quad $N(state, action)$ $\leftarrow$ $N(state, action) + n$
    \STATE \quad \textbf{end while} 
    \STATE \textbf{end function}
    \STATE
      \STATE {\bfseries Require:} Start from the root
      \WHILE {$episode < MAX_{-}episode$}
      \STATE $\text{Server}()$  
      \STATE $cur_{-}state$ $\leftarrow$ $root_{-}node$
      \STATE $i \leftarrow 0$
      \WHILE{$i < MAX_{-}tree_{-}depth$}
      \STATE $i \leftarrow i + 1$
      \STATE $next_{-}action$ $\leftarrow$ $\text{Selection}(cur_{-}state)$ \Commentl{Select an action based on Eq. \ref{ucb1_update}}
      \IF{$next_{-}state$ not in tree}
      \STATE $next_{-}state \leftarrow \text{Expansion}(next_{-}action)$
      \STATE $T_t \leftarrow \text{Simulation}_t(next_{-}state)$ for $t=0...k$ \Comment{$k$ is the number of simulations we run using the Meta-DNN}
      \STATE $TASK_{-}QUEUE.\text{push}(T_0)$
      \STATE $rollout_{-}from(T_0) \leftarrow next_{-}state$
      \STATE $\hat{q}(next_{-}state) \leftarrow \frac{1}{k} \sum_{i=1..k}Pred(T_i)$ \Comment{$Pred$ returns an accuracy predicted by the Meta-DNN}
      \STATE Backpropagation$(next_{-}state, \hat{q})$ \Commentl{Preemptive backpropagation to send $\hat{q}$}
      \ENDIF
      \ENDWHILE
      \ENDWHILE
    \end{algorithmic}
\end{algorithm}

\begin{table*}[h]
  \small
  \caption{The code of different layers}
  \label{app:layers_used}
  \centering
  \begin{tabular}{ l l l l l l l l}
    \toprule
          layers & code & layer & code & layer & code & layer & code \\
    \midrule
        3x3 avg pool & 1 & 3x3 max pool & 4 & 3x3 conv & 7 & 3x3 depth-separable conv & 10  \\
        5x5 avg pool & 2 & 5x5 max pool & 5 & 5x5 conv & 8 & 5x5 depth-separable conv & 11  \\
        7x7 avg pool & 3 & 7x7 max pool & 6 & identity & 9 & 7x7 depth-separable conv & 12  \\
    \bottomrule
    \label{layers_used}
  \end{tabular}
\end{table*}

\newpage
\section{Details of the State and Action Space}

\label{ap:state_space}
This section describes the state and action space for NASNet design space.

We constrain the state space to make the design problem manageable. The state space exponentially grows with the depth: a k layers linear network has $n^{k}$ architecture variations, where n is the number of layer types. We leverage the GPU DRAM size, our current computing resources and the design heuristics from leading DNNs, to propose the following constraints on the state space: 1) a branch has at most 1 layer; 2) a cell has at most 5 blocks; 3) the depth of blocks is limited to 2; 5) we use layers listed in TABLE.\ref{layers_used}.

Actions also preserve the constraints imposed on the state space. If the next state reaches out of the design boundary, the agent automatically removes the action
from the action set. For example, we exclude the "adding a new layer" action for a branch if it already has 1 layer. So, the action set is dynamically changing w.r.t states.


\section{Experiment Setup for Section \ref{architecture_search_eval} }
\label{ap:experiment_setup}

\subsection{Setup for searching networks on NASBench}

\textit{Regularized Evolution}: our implementation of Regularized Evolution~\cite{real2018regularized} is from \footnote{https://github.com/google-research/nasbench\\/blob/master/NASBench.ipynb}. The population size is set to 500 to find the global optimum, and the tournament size is 50. We only mutate the best architecture in the tournament set and replace the oldest individual in the population. We terminate the search once it finds the best architecture on NASBench.

\textit{AlphaX}: the only modification for AlphaX on NASBench is at the network training that queries an architecture performance instead of training. $c$ in Eq.~\ref{ucb1_update} is set to 2, and the number of predictions from meta-DNN is set to 10. The design and setup of meta-DNN are described in Sec.~\ref{sec:metadnn_design} and Sec.~\ref{ap:metadnn-architecture}.

\subsection{Setup for searching networks on CIFAR}
\label{ap:mcts_setup_cifar}
We used 8 NV-1080ti gpus for the open domain search. One GPU serves as the dedicated server, while the remaining gpus are clients for training the architectures sent from the server. 
The setup details are as follows: 
1) we early terminate the training at the 70th epoch, then rank networks to get top-10 architectures. We trained an additional 530 epochs on top-10 architectures to get their final accuracy; 
2) we applied cutout~\cite{zoph2017learning} during the training, and the cutout uses 1 crop of size $16\times16$; 
3) we applied cosine annealing learning rate schedule~\cite{cosine_restart} in the training, and the base learning rate is 0.025, the batch size is 96; 
4) we used the momentum optimizer with momentum rate set to 0.9 and L2 weight decay; 
5) we use dropout ratio schedule in the training. The droppath ratio is set to 0.3 and dense dropout ratio is set to 0.2 for searching procedure and applied \textit{ScheduleDropPath}~\cite{zoph2017learning} for the final training;
6) we use an auxiliary classifier located at 2/3 of depth of the network. The loss of the auxiliary classifier is weighted by 0.4~\cite{SzegedyVISW15}. 
7) the weights of architecture are initialized with Gaussian distribution with $\mu=0$ and $\sigma=0.01$. 
8) we randomly crop 32x32 patches from upsampled images of size $40\times40$, and apply random horizontal flips~\cite{zoph2017learning}.

\subsection{Setup for ImageNet}
\label{ap:setup_img}
The setup for training AlphaX-1 toward ImageNet are as follows: 
1) we construct the network for ImageNet with searched $RCell$ and $NCell$ according to Fig.~\ref{best-arch} in~\cite{zoph2017learning};
2) the input image size is $224\times224$;
3) our models for ImageNet use polynomial learning rate schedule, starting with 0.05 and decay through 200 epochs;
4) we use the momentum optimizer with momentum rate set to 0.9;
6) the dropout rate at the the final dense layer is 0.5; 
7) the weights of an architecture are initialized with Gaussian distribution with $\mu=0$ and $\sigma=0.01$.

\section{Experiment Setup for Section \ref{alg-comps-sec}}
\label{ap:comparison}

The state space consists of sequential ConvNet, e.g. AlexNet or VGG. 
The possible hyper-parameters of a convolution layer are stride $\in [1,2]$, filters $\in [32, 64]$, and kernels $\in [2,4]$, and the maximum depth of a ConvNet is set to 3.
The final layer of a ConvNet is always a dense layer followed by a softmax.

The possible actions are 1) adding a convolution layer; 2) choosing/changing an activation for a convolution layer; 3) changing a hyper-parameter of a convolution layer, e.g. filters from 32 to 64. At any state, the action set also consists of a special terminal action with a fixed probability that signals MCTS, random, and Q-learning to terminate the current search episode.

\textit{Random}: the agent randomly selects actions, and terminates the search once it hits a terminal action. Then the agent trains the network represented by the state where it hits the terminal to report accuracy.

\textit{Q-learning}: we used tabular Q-learning agent with $\epsilon$-greedy strategy. The learning rate is set to 0.2, and the discount factor is 1. We fixed $\epsilon$ to 0.2, and initialized the Q-value with 0.5.

\textit{Hill Climbing}: the agent trains every architecture in the child nodes, and greedily chooses the next child node in a fashion similar to PNAS~\cite{liu2017progressive}. We terminated the search once it sticks into a local optimum, and restarted the process at different random states.

\begin{figure}[t]
  \begin{center}
    \subfloat[][RNN predictor]{\includegraphics[height=1.4in]{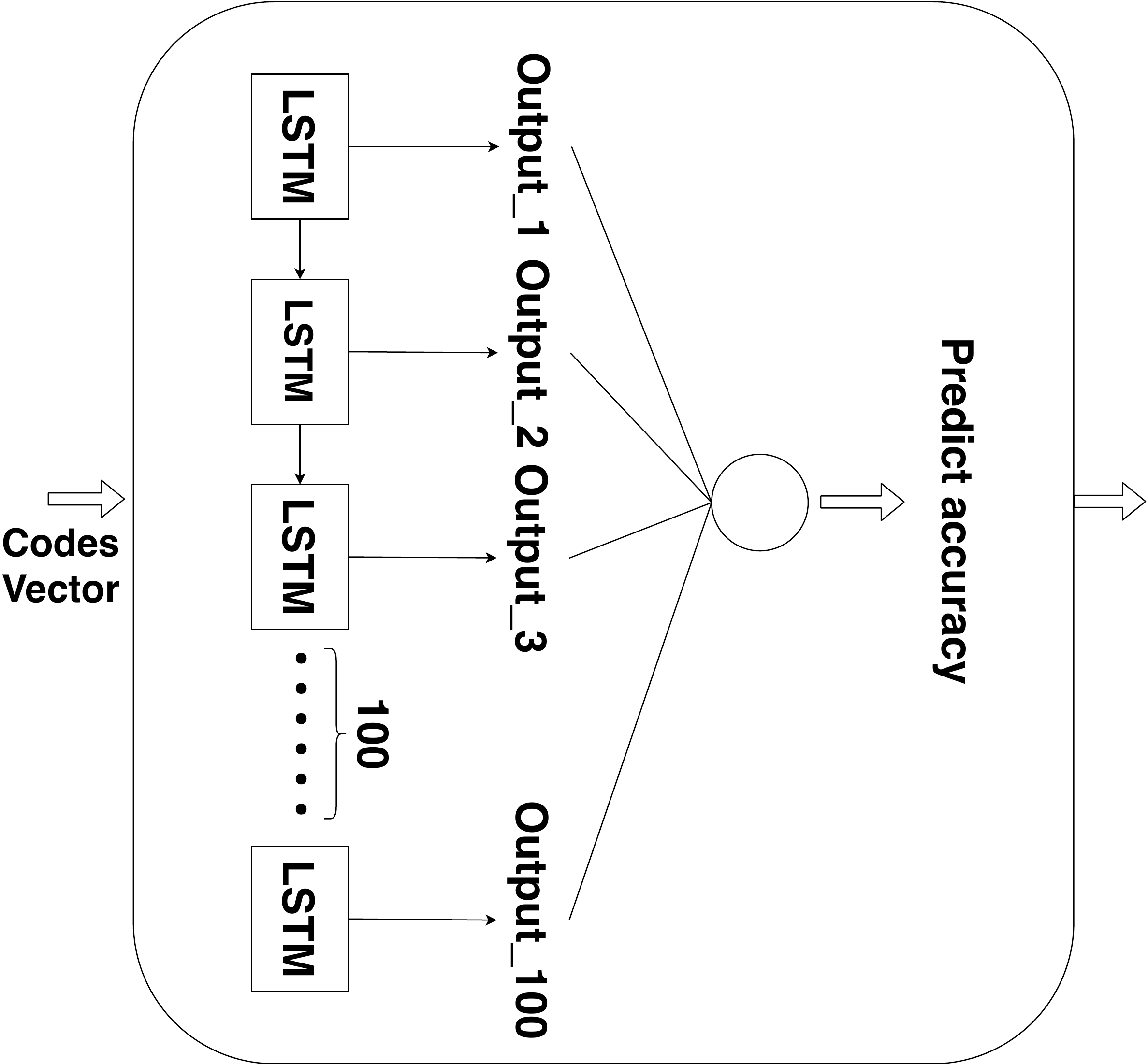}\label{meta_dnn_rnn}} \quad
    \subfloat[][MLP predictor]{\includegraphics[height=1.4in]{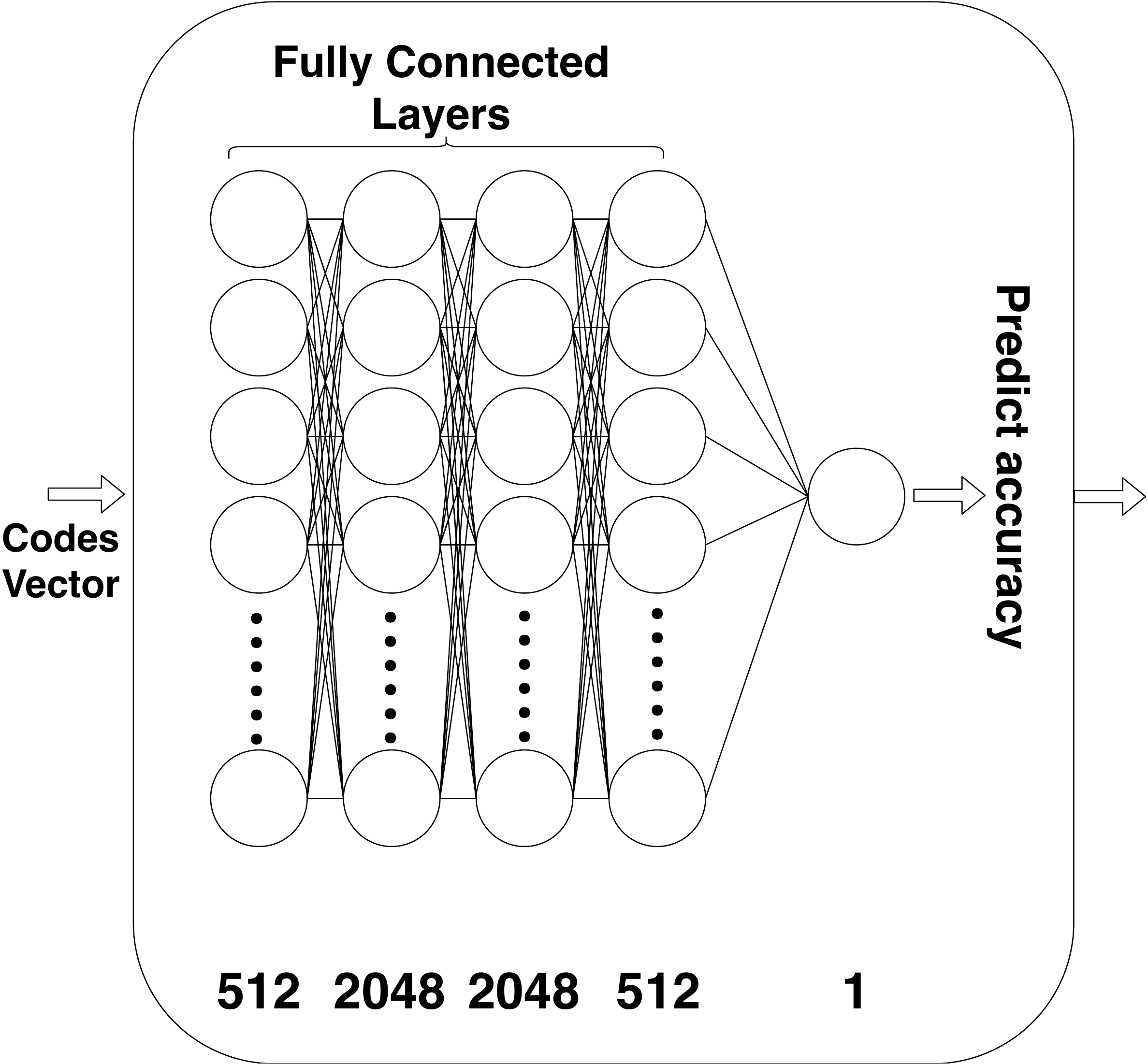}\label{meta_dnn_mlp}} 
  \end{center}
\caption{ \textbf{Architecture of metaDNN}: (a) the architecture of recurrent neural network style metaDNN (b) the architecture of multilayer perceptron style metaDNN. }
\label{meda_dnn}
\end{figure}

\begin{figure}[t]
\centering 
\includegraphics[height=2.5in, width=0.15\textwidth]{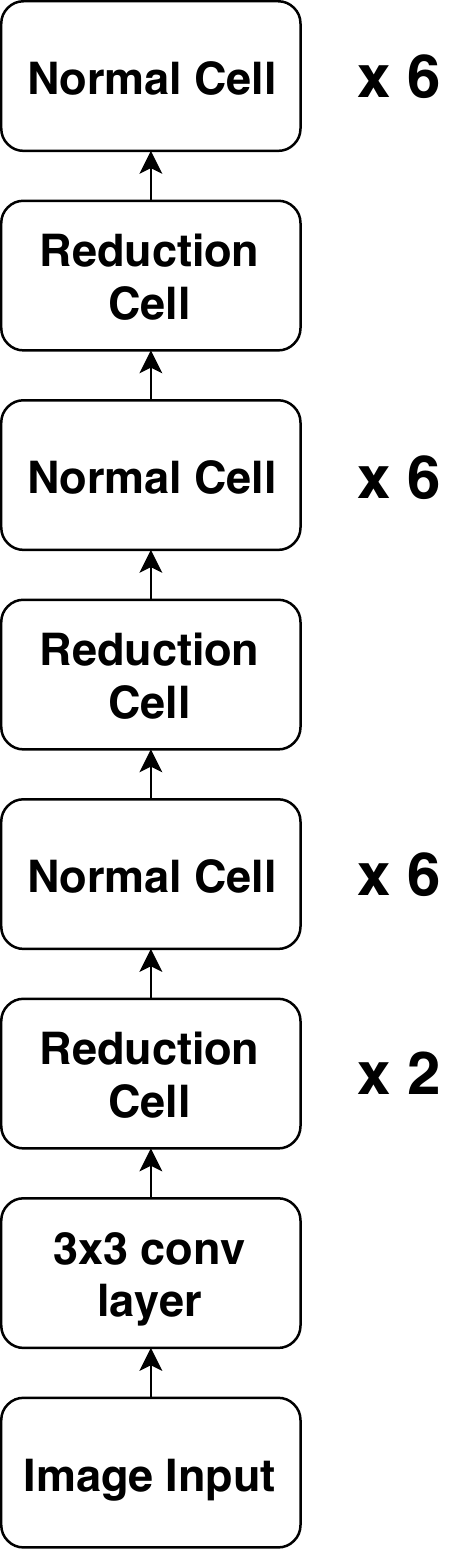}
\caption{Cell based architecture of AlphaX model}
\label{neural_style}
\end{figure}

\section{MetaDNN Designs and Setup}
\label{ap:metadnn-architecture}

\subsection{Setup for the RNN based meta-DNN}
The detailed configurations of LSTM based meta-DNN are as follows. The hidden state size is 100, and the embedding size is 100. 
The final LSTM hidden state goes through a fully-connected layer to get the final validation accuracy. Fig.\ref{meta_dnn_rnn} shows the architecture of the RNN predictor. 

The setup for the RNN training are: 1) the training lasts 20 epochs on currently collected samples; 2) the base learning rate is 0.00002 and the batch size is 128; 3) we use the Adam optimizer for the training; 4) we uniformly initialize the network in the range of [-0.1, 0.1].

\subsection{Setup for the MLP based meta-DNN}
The multi-stage model is an ensemble of multiple MLP models. The structure of an MLP is 512$\rightarrow$2048$\rightarrow$2048$\rightarrow$512$\rightarrow$1, as shown by Fig.\ref{meta_dnn_mlp}.
The setup for training a MLP are: 1) the training lasts 20 epochs on currently collected samples; 2) the learning rate is set to 0.00002 in an Adam optimizer and the batch size is 128;
4) we uniformly initialize the network in the range of [-0.1, 0.1].

\section{Setup for Vision Models}
\label{ap:setup_vision_models}
\subsection{Object detection}
We use AlphaX-1 model pre-trained on ImageNet dataset as the backbone. The training dataset is MSCOCO for object detection~\cite{LinMBHPRDZ14} which contains 90 classes of objects. Each image is scaled to $300\times300$ in RGB channels. We trained the model with 200k iterations with 0.04 initial learning rate and the batch size is set to 24. We applied the exponential learning rate decay schedule with the 0.95 decay factor. Our model uses momentum optimizer with momentum rate set to 0.9. We also use the L2 weight decay for training. We process each image with random horizontal flip and random crop~\cite{liu2016ssd}. We set the matched threshold to 0.5, which means only the probability of an object over 0.5 is effective to appear on the image. We use 8000 subsets of validation images in MSCOCO validation set and report the mean average precision (mAP) as computed with the standard COCO metric library~\cite{MSCOCO2014}. 

\subsection{Neural style}
We implement the neural style transfer application by replacing the VGG model to AlplaX-1 model~\cite{neural_style}. AlplaX-1 model is pre-trained on ImageNet dataset. In order to produce a nice result, we set the total 1000 iterations with 0.1 learning rate. We set 10 as the style weight which represents the extent of style reconstruction and 0.025 as the content weight which represents the extent of content reconstruction. We test different kinds of combinations of the outputs of different layers. Fig.~\ref{neural_style} shows the structure of AlphaX model, we found that for AlphaX-1 model, the best result can be generated by the concat layer of 13th normal cell as the feature for content reconstruction and the concat layer in first reduction cell as the feature for style reconstruction, the types of layers in each cell are shown in Fig.~\ref{best-arch}.

\subsection{Image captioning}
The training dataset of image captioning is MSCOCO~\cite{LinMBHPRDZ14}, a large-scale dataset for the object detection, segmentation, and captioning. Each image is scaled to $224\times224$ in RGB channels and subtract the channel means as the input to a AlphaX-1 model. For training AlphaX-1 model, We use the SGD optimizer with the 16 batch size and the initial learning rate is 2.0. We applied the exponential learning rate decay schedule with the 0.5 decay factor in every 8 epochs.

\end{document}